\documentclass[a4paper,11pt]{article}

\usepackage{amsmath}
\usepackage{amssymb}
\usepackage{amsthm}
\usepackage{graphicx}
\usepackage{authblk}
\usepackage{caption}
\usepackage{subcaption}
\usepackage[linesnumbered,ruled,vlined]{algorithm2e}
\usepackage{bm} 

\title{Machine Learning Enhanced Hankel Dynamic-Mode Decomposition}
\author[1,*]{Christopher W. Curtis}
\author[1,2]{D. Jay Alford-Lago}
\author[3, 4]{Erik Bollt}
\author[]{Andrew Tuma}
\affil[1]{Department of Mathematics and Statistics, San Diego State University, San Diego, CA, 92182, USA}
\affil[2]{Naval Information Warfare Center Pacific, San Diego, CA, 92152, USA}
\affil[3]{Department of Electrical and Computer Engineering, Clarkson University, 8 Clarkson Ave., Potsdam, NY, 13699, USA}
\affil[4]{Clarkson Center for Complex Systems Science, Clarkson University, 8 Clarkson Ave., Potsdam, NY, 13699, USA}
\affil[*]{Corresponding author: Christopher W. Curtis, ccurtis@sdsu.edu}

\date{}
\begin{document}
\maketitle
\section*{Abstract}
While the acquisition of time series has become more straightforward, developing dynamical models from time series is still a challenging and evolving problem domain.  Within the last several years, to address this problem, there has been a merging of machine learning tools with what is called the dynamic mode decomposition (DMD).  This general approach has been shown to be an especially promising avenue for accurate model development.  Building on this prior body of work, we develop a deep learning DMD based method which makes use of the fundamental insight of Takens' Embedding Theorem to build an adaptive learning scheme that better approximates higher dimensional and chaotic dynamics.  We call this method the Deep Learning Hankel DMD (DLHDMD). We likewise explore how our method learns mappings which tend, after successful training, to significantly change the mutual information between dimensions in the dynamics.  This appears to be a key feature in enhancing the DMD overall, and it should help provide further insight for developing other deep learning methods for time series analysis and model generation.  \\
\\
{\bf This work uses machine learning to develop an accurate method for generating models of chaotic dynamical systems using measurements alone.  A number of challenging examples are examined which show the broad utility of the method and point towards its potential impacts in advancing data analysis and modeling in the physical sciences.  Finally, we present quantitative studies of the information theoretic behavior of the machine learning tools used in our work, thereby allowing for a more detailed understanding of what can otherwise be an inscrutable method.}

\section{Introduction}

The incorporation of modern machine learning methodology into dynamical systems is creating an ever expanding array of techniques pushing the boundaries of what is possible with regards to describing and predicting nonlinear multi-dimensional time series.  Longstanding problems such as finding optimal Takens' embeddings \cite{fraser, sauer} now have powerful and novel deep learning based algorithmic approaches \cite{gilpin} which would not have been feasible even ten years ago.  Likewise, the field of equation free modeling using Koopman operator methods, broadly described by Dynamic Mode Decomposition (DMD), has seen several innovative deep learning based methods emerge over the last several years \cite{lusch, azencot, lago_dldmd} which have been shown to greatly expand the accuracy and flexibility of DMD based approaches.  There have also been related and significant advances in model identification and solving nonlinear partial differential equations via deep learning techniques \cite{champion2019data, kadierdan, karniadakis, li2022physicsinformed}.  

With this background in mind, in this work we focus on extending the methods in \cite{lago_dldmd} which were called Deep Learning DMD (DLDMD).  In that work, a relatively straightforward method merging auto-encoders with the extended DMD (EDMD) was developed.  This was done by using an encoder to embed dynamics in a sufficiently high enough dimensional space which then generated a sufficiently large enough space of observables for the EDMD to generate accurate linear models of the embedded dynamics.  Decoding then returned the embedded time series to the original variables in such a way as to guarantee the global stability of iterating the linear model to generate both reconstructions and forecasts of the dynamics.  The DLDMD was shown to be very effective in finding equation-free models which were able to both reconstruct and then forecast from data coming from planar dynamical systems.  

However, when chaotic time series from the Lorenz-63 system were examined, the performance of the DLDMD was found to degrade.  While this clearly makes the DLDMD approach limited in its scope, we note that the successful use of DMD based approaches to accurately reconstruct or forecast chaotic dynamics are not readily available.  Other methods such as HAVOK \cite{brunton_havok} or SINDy \cite{brunton_sindy} are more focused on the analysis of chaotic time series or the discovery of model equations which generate chaotic dynamics, though of course if one has an accurate model, then one should be able to generate accurate forecasts.  In this vein, there are also methods using reservoir computing (RC)  \cite{bollt2, bollt3}, though again, nonlinear models are essentially first learned and then used to generate forecasts.  However, both SINDy and RC rely on proposing libraries of terms to build models which are then fit (or learned from) data.  

While effective, such approaches do not allow for the spectral or modal analysis which has proven to be such an attractive and useful feature of DMD based methods.  Likewise, they require a number of user decisions about how to construct the analytic models used in later regressive fitting that amount to a guess and check approach to generating accurate reconstructions and forecasts.  Therefore in this work, using insights coming from the Takens' Embedding Theorem (TET) \cite{arbabi, gilpin}, we expand the DLDMD framework so as to make it accurate in both generating reconstructions and forecasts of chaotic time series.  This is done by first making the EDMD over embedded coordinates global as opposed to the local approach of \cite{lago_dldmd}; see also \cite{lusch, azencot}.  Second, we develop an adaptive Hankel matrix based ordering of the embedded coordinates which adds more expressive power for approximating dynamics to the deep learning framework.  To study our method, we use data generated by the Lorenz-63 and Rossler systems as well as twelve-dimensional projections of data from the Kuramoto--Sivashinksky (KS) equation.  In all of these cases, we show that by combining our proposed modifications to the DLDMD that we are able to generate far more accurate reconstructions and forecasts for chaotic systems than with DLDMD alone.  Moreover, we have built a method which still allows for the straightforward modal analysis which DMD affords and keeps user choices to a handful of real-valued hyperparamters while still producing results competitive with other approaches in the literature.    

Further, motivated by the classic information theory (IT) studies of the TET \cite{fraser}, as well as modern insights into the role that information plays in deep learning \cite{tishby1, calin}, we study how the fully trained encoder changes the information content of the dynamics coming from the Lorenz-63 and Rossler systems. For the Lorenz-63 system, the encoder tends to either slightly decrease the mutual information or cause strong phase shifts which decrease the coupling times across dimensions.  However, the characteristic timescales corresponding to lobe switching in the Lorenz `butterfly' are clearly seen to be preserved in the dynamics of the information for the Lorenz-63 system.  In contrast then, for the Rossler system, the slow/fast dichotomy in the dynamics seen in the original coordinates is made more uniform so that rapid transients in the information coupling are removed by the encoder.  Thus in either case, we see that the encoder generates significant differences in the information content between dimensions in the latent coordinates relative to the original ones, and that this strong change in information content is a critical feature in successful training.  

Of course, the present work is ultimately preliminary, and there are a number of important questions left to be resolved.  First, while we are able to easily display computed spectra, the affiliated global Koopman modes we find are not as straightforward to show.  We generate our results from random initial conditions, so the most effective means of constructing the global Koopman modes would be via radial-basis functions, but the implementation would be nontrivial due to the infamous ill-conditioning issues which can plague the approach \cite{fasshauer}. Second, there is a clear need for a comparison across SINDy, RC, and our DLHDMD methods.  In particular, the present work generates excellent reconstructions, but so far the predictive horizon is relatively short and difficult to increase.  How well other methods address this issue relative to their reconstruction and other diagnostic properties, and then how all of these methods compare in these several different ways is as yet unclear.  While acknowledging then the limitations of the present work, we defer addressing the above issues till later works where each of the above issues can be dealt with in the detail that is needed.  

The structure of this paper is as follows.  In Section 2, we provide an introduction to the Extended DMD and then explain the extensions we develop which are critical to the success of the present work.  In Section 3, we introduce the Hankel DMD and, incorporating the extensions introduced in Section 2, we show how well it does and does not perform on several examples.  Then in Section 4 we introduce the Deep Learning Hankel DMD and provide results on its performance as well as an analysis of how the mutual information changes in the latent variables.  Section 5 presents our results on mutual information.  Section 6 provides conclusion and discussion.
\section{Extended Dynamic Mode Decomposition}

To begin, we suppose that we have the data set $\left\{{\bf y}_{j}\right\}_{j=1}^{N_{T}+1}$ where
$$
{\bf y}_{j} = \varphi(t_{j};{\bf x}), ~ t_{j+1} = t_{j} + \delta t, ~ {\bf x}\in \mathbb{R}^{N_{s}}
$$
where $\delta t$ is the time step at which data is sampled and $\varphi(t;{\bf x})$ is a flow map such that $\varphi(t_{1},{\bf x})={\bf x}$.  From the flow map, we define the affiliated {\it Koopman operator} $\mathcal{K}^{t}$ such that for a given scalar observable $g({\bf x})$, one has 
\[
\mathcal{K}^{t}g({\bf x}) = g(\varphi(t,{\bf x})),
\]
so that the Koopman operator linearly tracks the evolution of the observable along the flow.  We likewise define the associated Hilbert space of {\it observables}, say $L_{2}\left(\mathbb{R}^{N_{s}},\mathbb{R},\mu\right)$, or more tersely as $L_{2}\left(\mathcal{O}\right)$, so that $g \in L_{2}\left(\mathcal{O}\right)$ if
$$
\int_{\mathbb{R}^{N_{s}}} \left|g({\bf x})\right|^{2} d\mu\left({\bf x}\right) < \infty,
$$
where $\mu$ is some appropriately chosen measure.  This makes the infinite-dimensional Koopman operator $\mathcal{K}^{t}$ a map such that  $
\mathcal{K}^{t}:L_{2}\left(\mathcal{O}\right)\rightarrow L_{2}\left(\mathcal{O}\right).$  

Following \cite{williams, williams2}, given our time snapshots $\left\{{\bf y}_{j}\right\}_{j=1}^{N_{T}+1}$, we suppose that any observable $g({\bf x})$ of interest lives in a finite-dimensional subspace $\mathcal{F}_{D}\subset L_{2}\left(\mathcal{O}\right)$ described by a given basis of observables $\left\{\psi_{l}\right\}_{l=1}^{N_{ob}}$ so that 
$$
g({\bf x}) = \sum_{l=1}^{N_{ob}}a_{l}\psi_{l}\left({\bf x}\right).
$$
Given this ansatz, we then suppose that  
\begin{align*}
\mathcal{K}^{\delta t}g({\bf x}) = &\sum_{l=1}^{N_{ob}}a_{l}\psi_{l}\left(\varphi\left(\delta t, {\bf x}\right) \right) \\
 = & \sum_{l=1}^{N_{ob}}\psi_{l}({\bf x})\left(\mathbf{K}^{T}_{a}{\bf a} \right)_{l} + r({\bf x};\mathbf{K}_{a})
\end{align*}
where $r({\bf x};\mathbf{K}_{a})$ is the associated error which results from the introduction of the finite-dimensional approximation of the Koopman operator represented by $\mathbf{K}_{a}$.  
We can then find ${\bf K}_{a}$ by solving the following minimization problem
\begin{align}
\mathbf{K}_{a} = & \mbox{arg min}_{\mathbf{K}} \left|r({\bf x};\mathbf{K})\right|^{2} \label{optproblem}\\
= & \mbox{arg min}_{\mathbf{K}}\sum_{j=1}^{N_{T}}\left|\sum_{l=1}^{N_{ob}}\left(a_{l}\psi_{l}({\bf y}_{j+1}) - \psi_{l}({\bf y}_{j})\left(\mathbf{K}^{T}{\bf a}\right)_{l}\right) \right|^{2}\nonumber \\
= & \mbox{arg min}_{\mathbf{K}} \sum_{j=1}^{N_{T}}\left|\left<\mathbf{\Psi}_{j+1} - \mathbf{K}\mathbf{\Psi}_{j} ,{\bf a}^{\ast}\right> \right|^{2}, \nonumber 
\end{align}
where ${\bf a}=(a_{1} \cdots a_{N_{ob}})^{T}$, $\mathbf{\mathbf{\Psi}}_{j} = \left(\psi_{1}({\bf y}_{j}) \cdots \psi_{N_{ob}}({\bf y}_{j}) \right)^{T}$, the inner product $\left<,\right>$ is the standard one over $\mathbb{C}^{N_{ob}}$, and the $\ast$ symbol denotes complex conjugation.  It is straightforward to show that an equivalent and easier to solve form of this optimization problem is given by 
\begin{equation}\label{eqn:dmd1}
    \mathbf{K}_{a} = \underset{\mathbf{K}}{\mathrm{argmin}}\left|\left|\mathbf{\Psi}_{+} - \mathbf{K}\mathbf{\Psi}_{-}\right|\right|_{F}^{2},
\end{equation}
where $\left|\left|\cdot\right|\right|_{F}$ is the Frobenius norm, and the $N_{ob}\times N_{T}$ matrices $\mathbf{\Psi}_{\pm}$ are given by 
\[
\mathbf{\Psi}_{-} = \left\{\mathbf{\Psi}_{1}~ \mathbf{\Psi}_{2}~ \cdots ~\mathbf{\Psi}_{N_{T}} \right\}, \quad \mathbf{\Psi}_{+} = \left\{\mathbf{\Psi}_{2} ~\mathbf{\Psi}_{3} ~\cdots ~\mathbf{\Psi}_{N_{T}+1} \right\}.
\]

In practice, we solve this equation using the Singular-Value Decomposition (SVD) of $\mathbf{\Psi}_{-}$ so that
\begin{align*}
    \mathbf{\Psi}_{-} = \mathbf{U} \mathbf{\Sigma} \mathbf{W}^\dagger.
\end{align*}
This then gives us 
\begin{equation*} \label{eqn:dmd2}
    \mathbf{K}_{a} = \mathbf{\Psi}_{+} \mathbf{W} \mathbf{\Sigma}^{-1} \mathbf{U}^{\dagger}.
\end{equation*} 
To complete the algorithm, after diagonalizing $\mathbf{K}_{a}$ so that 
\begin{equation}
\mathbf{K}_{a} = \mathbf{V} \mathbf{L} \mathbf{V}^{-1}, ~\mathbf{L}_{ll} = \ell_{l}, 
\label{diagequation}
\end{equation}
then one can show that the Koopman eigenfunctions $\phi_{l}({\bf y}_{j})$ are found via the equations
\begin{equation}
\mathbf{\Phi}_{\pm} = {\bf V}^{-1}\mathbf{\Psi}_{\pm}.
\label{phimat}
\end{equation}
From here, one can, starting from the initial conditions, approximate the dynamics via the reconstruction formula  
\begin{equation}
{\bf y}(t;{\bf x})\approx \sum_{l=1}^{N_{ob}}{\bf k}_{l}e^{t\lambda_{l}}\phi_{l}({\bf x}),
\label{windowformula}
\end{equation}
where $\lambda_{l} = \ln(\ell_{l})/\delta t$ and the {\it Koopman modes} ${\bf k}_{l}\in \mathbb{C}^{N_{s}}$ solve the initial-value problem 
\[
{\bf x} = \sum_{l=1}^{N_{ob}}{\bf k}_{l} \phi_{l}({\bf x}).  
\]
Again, in matrix/vector notation, keeping in mind that ${\bf x}\in \mathbb{R}^{N_{s}}$ and that in general $N_{s}\neq N_{ob}$, we have 
\[
{\bf x} = {\bf K}_{M}\begin{pmatrix}\phi_{1}({\bf x})\\ \vdots \\ \phi_{N_{ob}}({\bf x}) \end{pmatrix}
\]
where ${\bf K}_{M}$ is the $N_{s}\times N_{ob}$ matrix whose columns are the Koopman modes ${\bf k}_{j}$.  As can be seen then, generically, one can only find the Koopman modes through least-squares solutions of the non-square problem.  In this regard, one would do well to have information from as many initial conditions as possible to over-determine the problem.  

\subsection{Global EDMD}

To wit, if we had a collection of initial conditions $\left\{{\bf x}_{k}\right\}_{k=1}^{N_{C}}$ with corresponding path data $\left\{{\bf y}_{j,k}\right\}_{j,k=1}^{N_{T}+1,N_{C}}$, we can extend the optimization problem in Equation \eqref{optproblem} to be 
\[
\mathbf{K}_{a} =  \mbox{arg min}_{\mathbf{K}} \sum_{k=1}^{N_{C}} \left|r({\bf x}_{k};\mathbf{K})\right|^{2},
\]
so that now the problem of finding $\mathbf{K}_{a}$ is no longer strictly localized to a particular path labeled by the initial condition ${\bf x}$.  Following the same optimization argument as above leads one to concatenate across observables column wise when generating the $\mathbf{\Psi}_{\pm}$ matrices so that  
\[
\mathbf{\Psi}_{-} = \left\{\mathbf{\Psi}_{1,1}~ \mathbf{\Psi}_{2,1}~ \cdots ~\mathbf{\Psi}_{N_{T},1} ~ \cdots \mathbf{\Psi}_{1,N_{C}}~ \mathbf{\Psi}_{2,,N_{C}}~ \cdots ~\mathbf{\Psi}_{N_{T},N_{C}}\right\}
\]
where
\[
\mathbf{\Psi}_{j,k} = \left(\psi_{1}(\varphi(t_{j}; {\bf x}_{k})) \cdots \psi_{N_{ob}}(\varphi(t_{j}; {\bf x}_{k})) \right)^{T}
\]
The matrix $\mathbf{\Psi}_{+}$ is defined similarly.  Using then the EDMD algorithm outlined above, we arrive at the following matrix problem for determining ${\bf K}_{M}$
\[
{\bf X} = {\bf K}_{M} \mathbf{\Phi}_{0}
\]
where
\[
{\bf X} = \left({\bf x}_{1} \cdots {\bf x}_{N_{C}} \right), ~ 
\mathbf{\Phi}_{0} = \begin{pmatrix} 
\phi_{1}({\bf x}_{1}) & \cdots & \phi_{1}({\bf x}_{N_{C}}) \\
\phi_{2}({\bf x}_{1}) & \cdots & \phi_{2}({\bf x}_{N_{C}}) \\
\vdots & \vdots & \vdots \\ 
\phi_{N_{ob}}({\bf x}_{1}) & \cdots & \phi_{N_{ob}}({\bf x}_{N_{C}}).
\end{pmatrix}
\]

Likewise, given that Equation \eqref{phimat} gives us time series of the Koopman eigenfunctions, which necessarily must satisfy, assuming sufficient accuracy of the approximation implied by Equation \eqref{diagequation}, the identity
\[
\phi_{l}\left(\varphi(t_{j};{\bf x}_{k})\right) = \mathcal{K}^{j}\phi_{l}({\bf x}_{k}) = \ell_{l}^{j}\phi_{l}\left({\bf x}_{k}\right),
\]
we can generalize Equation \eqref{windowformula} via the model
\begin{align}
{\bf Y}_{N_{st}} \approx & {\bf K}_{M}{\bf L}^{N_{st}}\mathbf{\Phi}_{-}, \nonumber \\
\approx &\bar{{\bf K}}_{M}{\bf K}_{a}^{N_{st}}\mathbf{\Psi}_{-} \label{genwindowequation}
\end{align}
where $N_{st} \in \mathbb{N} \cup \left\{0\right\}$, $\bar{{\bf K}}_{M} = {\bf K}_{M}{\bf V}$, and 
\[
{\bf Y}_{N_{st}} \approx \left\{{\bf y}_{N_{st},1} \cdots {\bf y}_{N_{st}+N_{T},1} ~ \cdots ~   {\bf y}_{N_{st},N_{C}}\cdots {\bf y}_{N_{st}+N_{T},N_{C}}\right\},
\]
which generates a reconstruction of the data for time steps $N_{st}\leq j \leq N_{T}+1$ and a forecast for steps with index $N_{T}+1\leq j \leq N_{T}+N_{st}$.   Using this formula allows for far greater flexibility in employing the EDMD since we can control how many times steps for which we wish to generate reconstructions.  This is in contrast to generating forecasts through the iteration of the diagonal matrix ${\bf L}$, which is a process that is generally sensitive to small variations in the position of the eigenvalues $\ell_{l}$, especially for those near the unit circle in the complex plane.  We will make great use of this generalization in the later sections of this paper.  We also note that in order to avoid introducing complex values, which can cause significant complications when using current machine learning libraries, we will do all later computations in terms of $\bar{{\bf K}}_{M}$ and ${\bf K}_{a}$ as in Equation \eqref{genwindowequation}.

\section{Hankel DMD}

When implementing EDMD, the most natural observables are the projections along the canonical Cartesian axes, i.e. 
\[
\psi_{l}({\bf x}) = x_{l}, ~ l=1,\cdots, N_{s}.
\]
If we stick to this space of observables, the EDMD method reduces to the standard DMD method.  Thus the idea with EDMD is to include more nonlinear observables to hopefully represent a richer subspace of dynamics and thereby make the approximation of corresponding Koopman operator more accurate and sophisticated.

With this in mind, \cite{arbabi} built upon the classic idea of Takens embeddings \cite{takens} and explored using affiliated Hankel matrices to generate natural spaces of observables for EDMD, and approach we describe as Hankel DMD (HDMD).  Also of note in this direction is the HAVOK method developed in \cite{brunton_havok}, though in some ways HAVOK is more akin to the {\it embedology} methods explored in such classic works as \cite{broomhead, sauer}.  

HDMD thus begins with an affiliated scalar measurement of our time series, say $\left\{g({\bf y}_{j})\right\}_{j=1}^{N_{T}+1}$.  From this, by introducing a {\it window} size $N_{w}$ one builds the affiliated Hankel matrix ${\bf H}_{g}\left({\bf x}\right)$ where 
\[
{\bf H}_{g}\left({\bf x}\right) = \begin{pmatrix}
g({\bf y}_{1}) & \cdots & g({\bf y}_{N_{w}})\\
g({\bf y}_{2}) & \cdots & g({\bf y}_{N_{w+1}})\\
\vdots & \vdots & \vdots \\
g({\bf y}_{N_{ob}}) & \cdots & g({\bf y}_{N_{T}+1})
\end{pmatrix}.
\]
where the number of observables $N_{ob}= N_{T}+1-(N_{w}-1)$.

What one sees then is that each row of ${\bf H}_{g}({\bf x})$ is some iteration of the Koopman operator $\mathcal{K}^{\delta t}$.  From here then, each row of $N_{w}$ time steps is defined to be its own separate observable $\psi_{l}({\bf x})$, i.e. 
\[
\psi_{l}({\bf x}) = \mathcal{K}^{l \delta t}g({\bf x}), ~ l=1,\cdots,N_{ob}.
\]
One then proceeds as above with the EDMD algorithm, where we emphasize that $N_{T}$ is replaced by $N_{w}-1$.  This is an interesting feature, or arguably limitation, of the HDMD method in which we generate matrices $\mathbf{\Phi}_{\pm}$ (see Equation \eqref{phimat}) up to the time index $N_{w}-1\leq N_{T}$.   Thus later times are used to build approximations at prior times.  This makes the issue of forecasting data more difficult since one must iterate the EDMD results, as is done via Equation \eqref{genwindowequation}, from time index $N_{w}-1$ up to $N_{T}$ to reconstruct the original data that was used in the first place.  Throughout the remainder of the paper then, we take care to distinguish between {\it iterated reconstructions} and actual {\it forecasts} which make novel predictions beyond the given data.  

Finishing our explanation of HDMD, if one has data along multiple initial conditions, say $\left\{{\bf x}_{k}\right\}_{k=1}^{N_{C}}$, we can extend the above algorithm by concatenating Hankel matrices so that we perform EDMD on the combined matrix ${\bf H}_{C}$ so that 
\[
{\bf H}_{C} = \left({\bf H}_{g}\left({\bf x}_{1}\right) ~ \cdots ~ {\bf H}_{g}\left({\bf x}_{N_{C}}\right) \right)
\]
The inclusion of other observables can be done in a similar fashion.
  
\subsection{Results for HDMD}
The ultimate promise of the HDMD is that it should facilitate an adaptable implementation of the EDMD framework which allows for the number of observables to simply be adjusted by the window size.  To see this, in all of the following results we let $t_{f}=20$, $dt=.05$, and we use $N_{C}=128$ random initial conditions which are then stacked together.  For HDMD, observables along each dimension of the dynamical system are used so that 
\[
N_{ob} = N_{s}\bar{N}_{ob}, ~ \bar{N}_{ob} = N_{T}+1-(N_{w}-1)
\]
  Reconstructions and forecasts are generated using Equation \eqref{genwindowequation} for $N_{st}=20$, which for a time step of $dt=.05$, means forecasts are produced up to a unit of non-dimensional time.  We note though that the choice of $\bar{N}_{ob}$ sets the window size value $N_{w}=N_{T}+1-(\bar{N}_{ob}-1)$, so that instead of using EDMD on data from $0\leq t\leq t_{f}$, we now use data from $0\leq t \leq t_{f,w}$ where $t_{f,w} = N_{w}\delta t$.  

\begin{figure}[!h]
\begin{tabular}{cc}
\includegraphics[width=.45\textwidth]{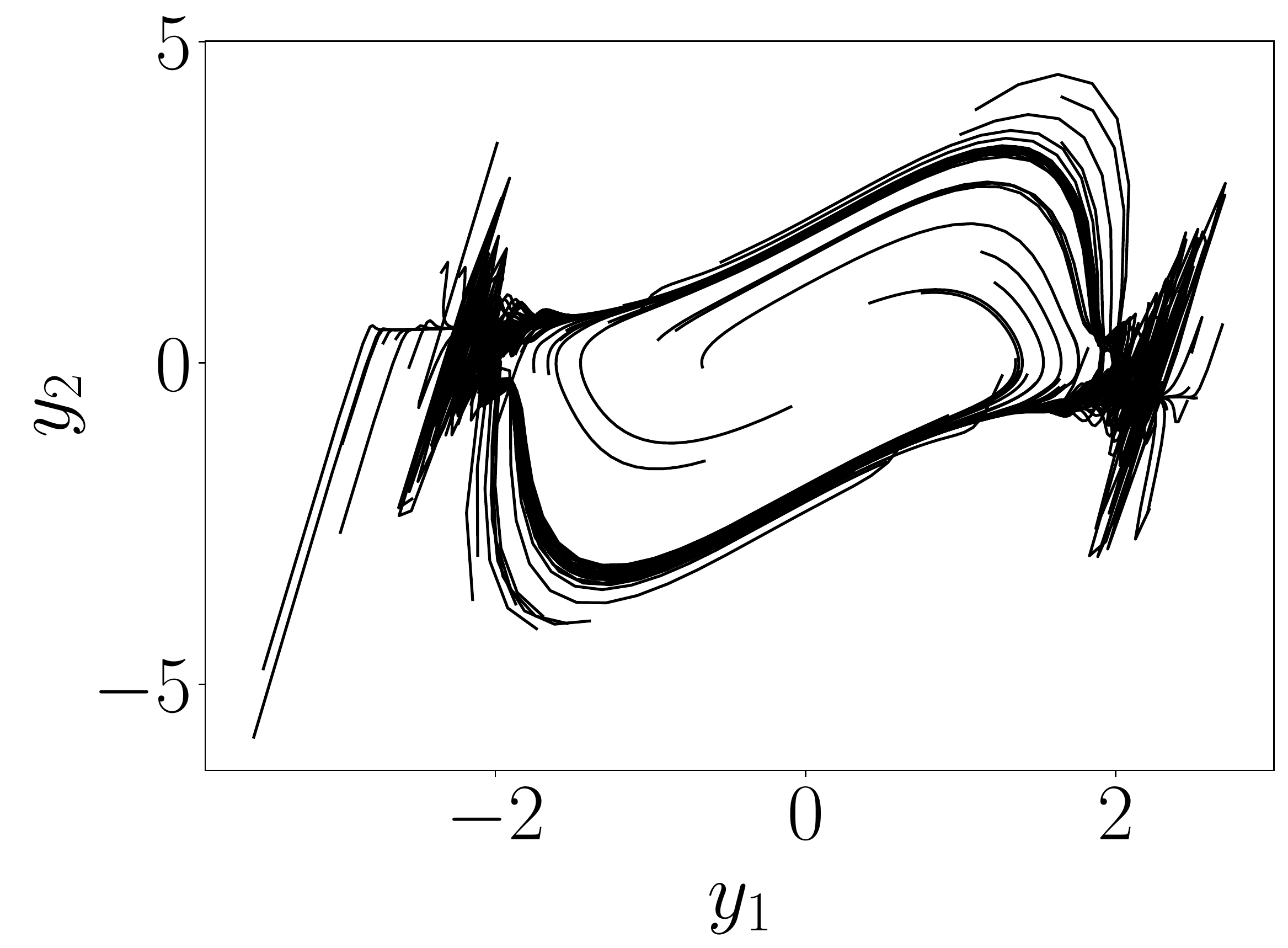} & \includegraphics[width=.46\textwidth]{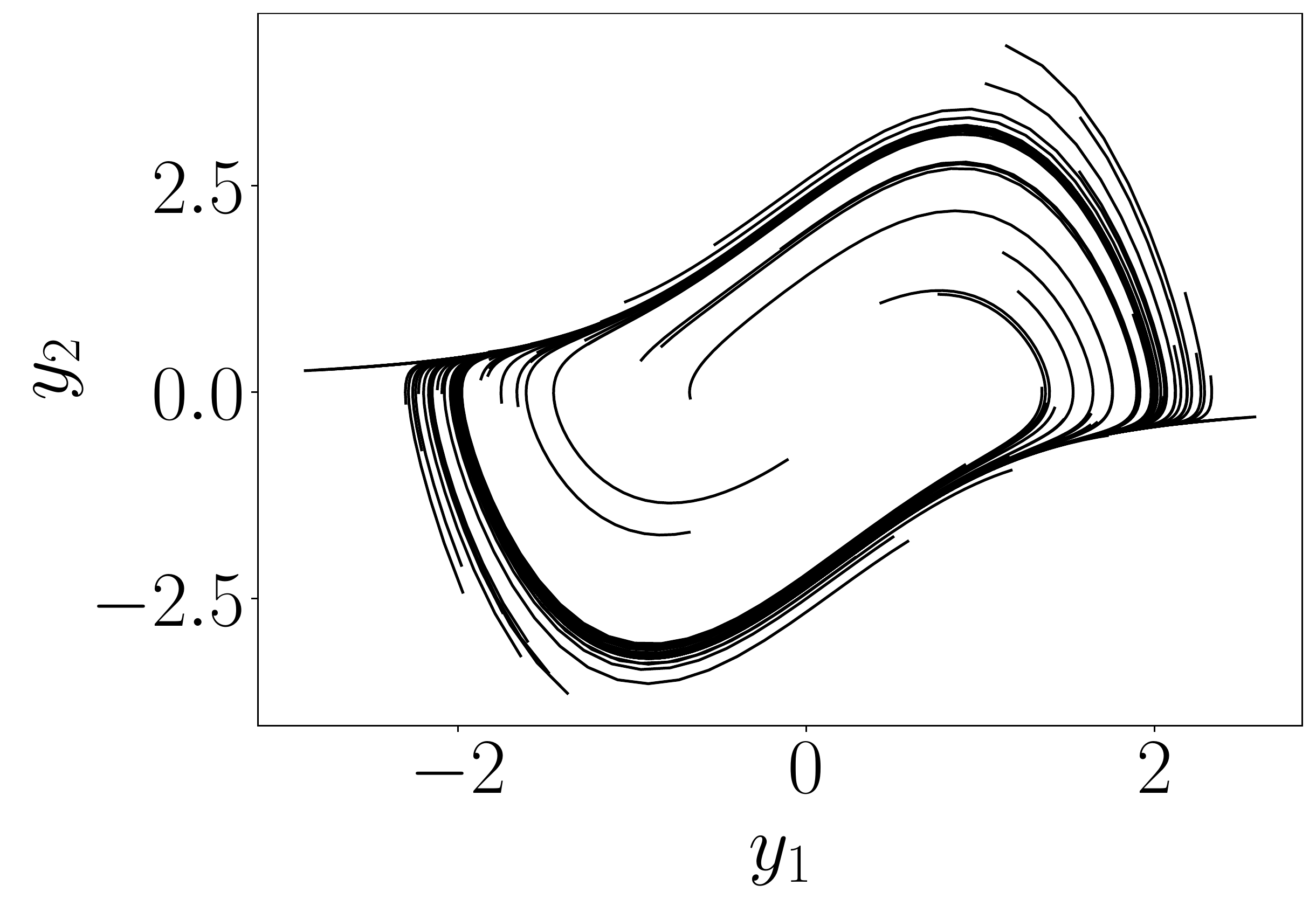}
\end{tabular}
\caption{HDMD results for the Van der Pol equation with $N_{st}=20$ and $\bar{N}_{obs}=10$ (left) and $\bar{N}_{obs}=20$ (right), so that the reconstruction is generated for $1\leq t\leq t_{f,w}$ and iterated reconstruction and forecasting is done for times $t_{f,w}\leq t \leq t_{f,w}+1$ where $t_{f,w}=19.5$ for $\bar{N}_{obs}=10$ and $t_{f,w}=19$ for $\bar{N}_{obs}=20$.  For $\bar{N}_{ob}=20$ the maximum error over trajectories is $.0214\%$ while for $\bar{N}_{ob}=10$ it is $25.9 \%$.  The enhanced accuracy comes at the expense of generating novel forecasts of the time series.  As can be seen, relative to the choice of $N_{st}$, doubling the number of observables greatly enhances the accuracy of the reconstructions and forecast.}
\label{fig:hankelvan}
\end{figure}

If we then take data from the Van der Pol oscillator, where
\[
\dot{y}_{1} = y_{2}, ~ \dot{y}_{2} = -y_{1} + \mu(1-y_{1}^{2})y_{2},  ~ \mu =1.5, 
\]
we find, as seen in Figure \ref{fig:hankelvan}, that $\bar{N}_{ob}=10$ does not produce especially accurate reconstructions or forecasts.  By increasing $\bar{N}_{ob}$ to $20$ though, we are able to generate far more accurate results, though at the cost of being able to forecast beyond the given time series.  We further note that by fixing $\bar{N}_{ob}=20$ and letting $N_{st}=30$, we get essentially the same degree of degradation in the forecast as when we chose $\bar{N}_{ob}=10$ and $N_{st}=20$.  

\begin{figure}[h]
\begin{tabular}{cc}
\includegraphics[width=2.25in, height=1.75in]{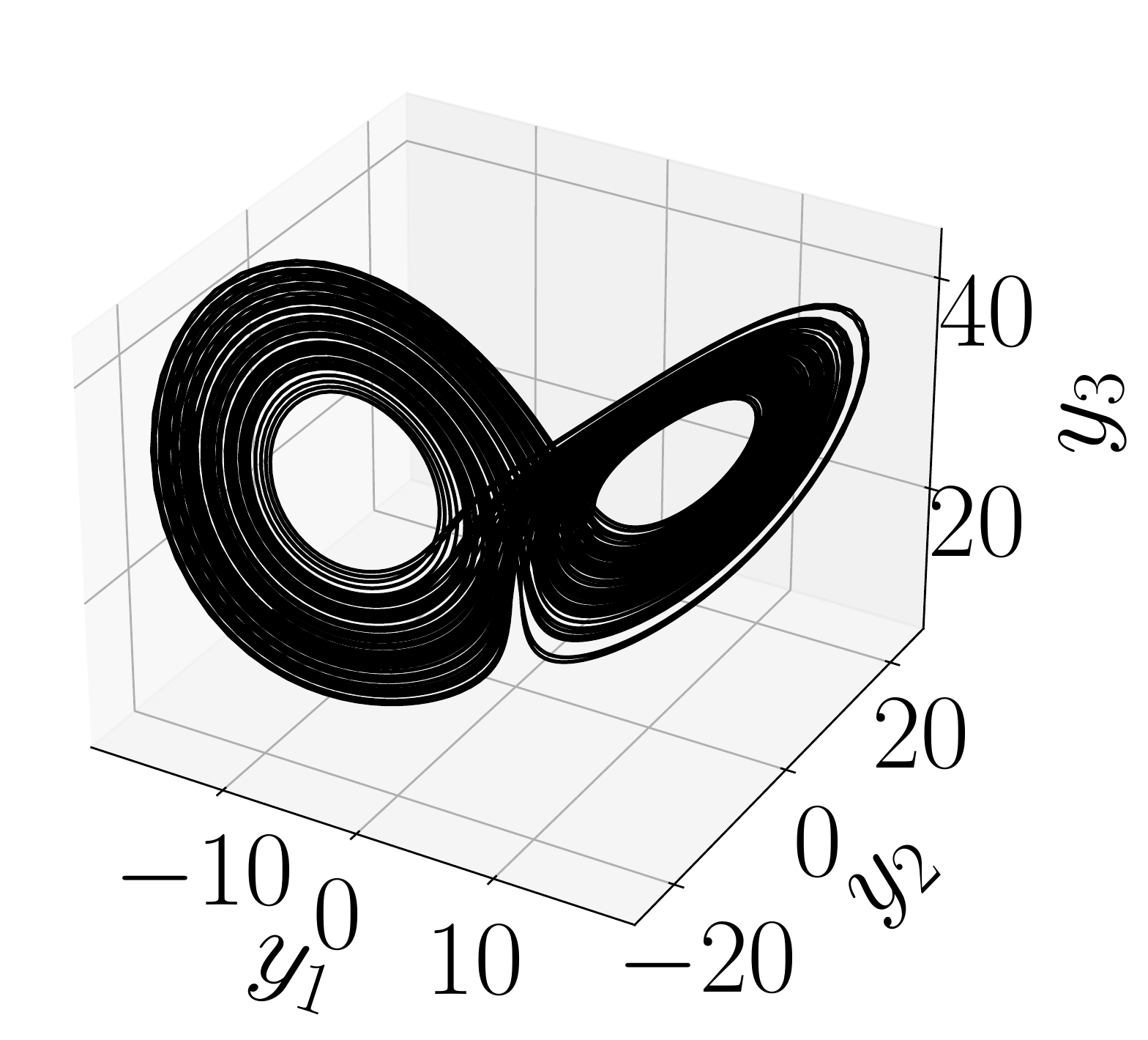}  & \\
\includegraphics[width=2.25in, height=1.75in]{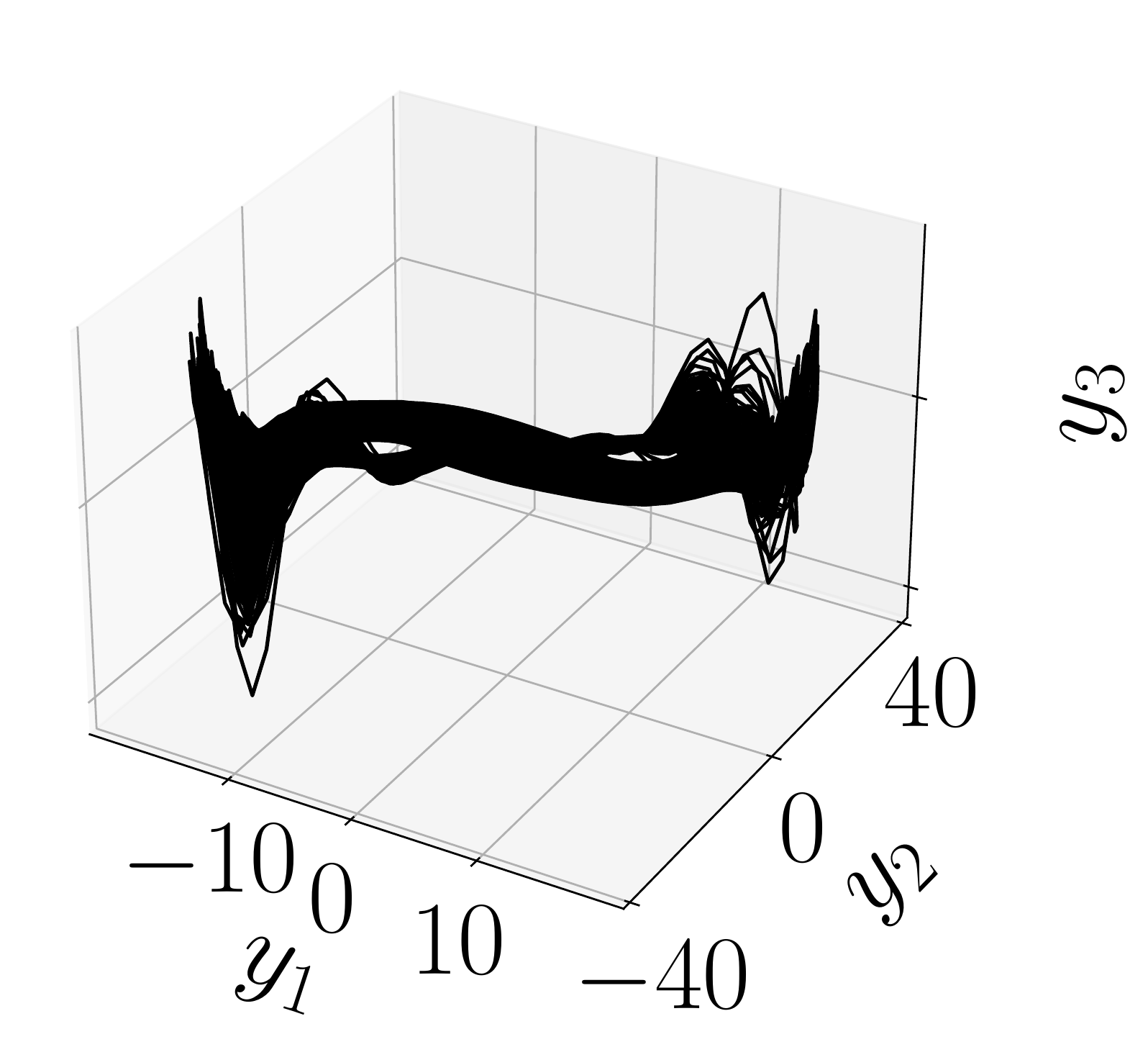} & \includegraphics[width=2.25in, height=1.75in]{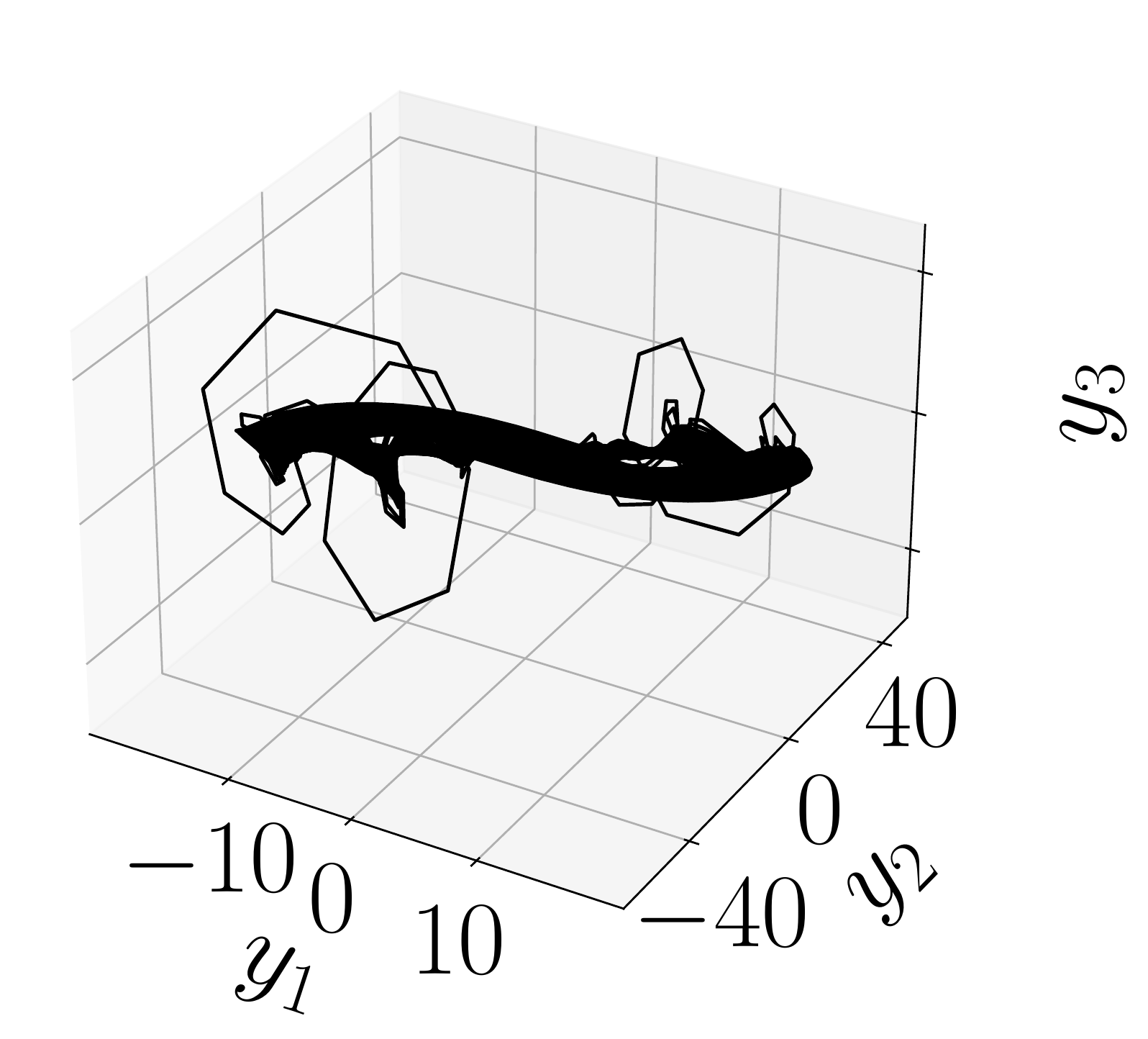} 
\end{tabular}
\caption{In the top row, the Lorenz Butterfly.  In the bottom row, the HDMD results for the Lorenz system with $N_{st}=20$, so that the reconstruction is generated for times $1\leq t\leq t_{f,w}$ and iterated reconstruction and forecasting is done for times $t_{f,w}\leq t \leq t_{f,w}+1$ where $t_{f,w}=19.5$ for $\bar{N}_{ob}=10$ (left) and $t_{f,w}=19$ for $\bar{N}_{ob}=20$ (right).  As can be seen, doubling the number of observables does little to enhance the accuracy of the reconstructions, and the errors in the bottom row plots are well over $100\%$.}
\label{fig:lorenzapprox}
\end{figure}

In contrast to these results, we find that the Lorenz Equations
\begin{align}
\dot{y}_{1} = & \sigma(y_{2}-y_{1}),\nonumber \\ 
\dot{y}_{2} = & \rho y_{1} - y_{2} - y_{1}y_{3},\nonumber \\ 
\dot{y}_{3} = & -by_{3} + y_{1}y_{2},  \label{lorenzequation}
\end{align}
where
\[
\sigma =  10, ~ \rho = 28, ~ b = \frac{8}{3},
\]
provide a case in which the HDMD is not able to adequately capture the dynamics for any reasonable choices of $\bar{N}_{ob}$.  This is not necessarily surprising since for the parameter choices made, we know that the dynamics traces out the famous {Butterfly} strange attractor as seen in the top row of Figure \ref{fig:lorenzapprox}.  Given that we are trying to approximate dynamics on a strange attractor, we would reasonably anticipate the HDMD to struggle.  

However, as seen in the bottom row of Figure \ref{fig:lorenzapprox}, the method essentially fails completely for parameter choices identical to those used above for the Van der Pol oscillator.  That all said, further adaptation of the HDMD method might produce more desirable results.  We will see how to realize this through the use of neural networks in the following section.  

\section{Deep Learning HDMD}

To improve the HDMD such that it is able to deal with chaotic systems such as the Lorenz equation, we now turn to and adapt the framework of the deep learning DMD (DLDMD) developed in \cite{lago_dldmd}.  Our deep learning enhanced HDMD begins with an autoencoder composed of neural networks $\mathcal{E}$ (the encoder) and $\mathcal{D}$ (the decoder) such that
\[
\mathcal{E}: \mathbb{R}^{N_{s}} \rightarrow \mathbb{R}^{N_{e}}, ~ \mathcal{D}: \mathbb{R}^{N_{e}} \rightarrow \mathbb{R}^{N_{s}}, 
\]
and such that our auto-encoder is a near identity, i.e. 
\[
 \tilde{{\bf y}} = \mathcal{E}\left({\bf y}\right), ~ \mathcal{D} \left(\tilde{{\bf y}} \right) \approx {\bf y}.
\]
Note, we call the encoded coordinates {\it latent variables} or {\it latent dimensions} in line with the larger literature on machine learning.  Motivated by our results in \cite{lago_dldmd}, we take $N_{e}\geq N_{s}$.  We then have in the later EDMD step that $N_{ob} = N_{e}\bar{N}_{ob}$, so that larger values of $N_{e}$ correspond to more observables to represent the dynamics through an approximation to the Koopman operator.  

The encoded coordinates should represent a set of observables which should enhance the overall accuracy of HDMD approximations of the dynamics.  To train for this, after making reasonable choices for how to initialize the weights of the auto-encoder, and fixing a choice for $N_{st}$, given the training data $\left\{{\bf y}_{j,k}\right\}_{j,k=1}^{N_{T}+1, N_{C}}$, after shuffling into $\lfloor N_{C}/N_{B}\rfloor$ batches of batch size $N_{B}$, over a given batch ${\bf Y} = \left\{{\bf y}_{j,k}\right\}_{j,k=1}^{N_{T}+1, N_{B}}$, we use the following loss function
\[
\mathcal{L}_{tot} = \mathcal{L}_{recon} + \mathcal{L}_{pred} + \mathcal{L}_{dmd} + \alpha \mathcal{L}_{reg}
\]
where
\begin{align*}
\mathcal{L}_{recon} = & \left[\left[\left|\left|{\bf Y} -  \mathcal{D}\circ \mathcal{E}\left({\bf Y} \right)\right|\right|_{2}\right]_{k=1}^{N_{B}}\right]_{j=1}^{ N_{T}+1}, \\
\mathcal{L}_{dmd} = & \left[\left[\left|\left|\tilde{{\bf Y}} -  \bar{{\bf K}}_{M}{\bf K}_{a}^{N_{st}}\mathbf{\Psi}_{-} \right|\right|_{2}\right]_{k=1}^{N_{B}}\right]_{j=N_{st}}^{N_{w}-1},\\
\mathcal{L}_{pred} = &  \left[\left[ \left|\left|{\bf Y} -  \mathcal{D} \left( \bar{{\bf K}}_{M}{\bf K}_{a}^{N_{st}}\mathbf{\Psi}_{-}\right)\right|\right|_{2}\right]_{k=1}^{N_{B}}\right]_{j=N_{st}}^{N_{w}-1},
\end{align*}
with $[\cdot]_{k=1}^{N_{B}}$ denoting averaging over a given batch, and $[\cdot]_{j=j_{1}}^{j_{2}}$ denoting averaging over the $j_{1}$ to $j_{2}$ timesteps.  We also emphasize here that the matrices $\bar{{\bf K}}_{M}$, ${\bf K}_{a}$, and $\mathbf{\Psi}_{-}$ are defined globally over a given batch of trajectories.  This is in contrast to the methods in \cite{lago_dldmd} or \cite{lusch} which computes equivalent matrices over each trajectory.  Arguably then, our approach is learning more global representations of the Koopman operator, the consequences of which we will explore in future work, though see the Appendix for a brief exploration on this point.  

\subsubsection{Updating $\bar{N}_{ob}$}
One of the most compelling reasons to use Hankel-DMD in our learning algorithm is that it easily allows us to change the model without having to change the underlying neural networks represented by $\mathcal{E}$ and $\mathcal{D}$.  We note that this is in marked contrast to the method of \cite{lago_dldmd}, which while allowing for varying models by changing the dimensionality of the range of $\mathcal{E}$, then must fix that output dimension for the length of training.  This can make finding good models a challenging task.  Thus, we anticipate by allowing $\bar{N}_{ob}$ to change that we can develop far more flexible models which are ultimately easier to train and use.  

To update $\bar{N}_{ob}$, we first train the networks $\mathcal{E}$ and $\mathcal{D}$ by optimizing $\mathcal{L}_{tot}$ over a given epoch.  Then, fixing the weights in the networks, we take a subset of the training data and compute $\mathcal{L}_{tot}(\bar{N}_{ob})$.  We then compute $\mathcal{L}_{tot}(\bar{N}_{ob}\pm1)$ and compare to $\mathcal{L}_{tot}(\bar{N}_{ob})$.  If either of the variant models gives a lower value of $\mathcal{L}_{tot}$ and causes a relative change greater than some chosen tolerance $f_{r}$, then we update the model by changing $\bar{N}_{ob}$.  Note, the choice of the range of values of $\bar{N}_{ob}$ comes from trial and error with a particular emphasis on computational efficiency and cost.  Likewise, introducing the relative change threshold $f_{r}$ was found to be useful in ignoring what amounted to spurious changes in the model parameters.  We collect the details of our learning method in Algorithm \ref{dlhdmd}, which we call the Deep Learning HDMD (DLHDMD).
\begin{algorithm}
\SetKwData{Left}{left}\SetKwData{This}{this}\SetKwData{Up}{up} \SetKwFunction{Union}{Union}\SetKwFunction{WindowUpdate}{Window Update} \SetKwInOut{Input}{input}\SetKwInOut{Output}{output}
\KwData{Choose parameters $N_{e}$, $N_{C}$, $N_{B}$, $\alpha$, $f_{r}$, $N_{st}$}
\KwData{Choose initial value of $\bar{N}_{ob}$}
\For{$l\leftarrow 1$ \KwTo $E_{max}$ }
{
Shuffle $N_{C}$ trajectories into $\lfloor N_{C}/N_{B}\rfloor$ batches each of size $N_{B}$\;
\For{$m \leftarrow 1$ \KwTo $\lfloor N_{C}/N_{B}\rfloor$ }
{ 
$\tilde{{\bf Y}} \leftarrow \mathcal{E}({\bf Y})$ \;
$\bar{{\bf K}}_{M},~{\bf K}_{a}, ~\mathbf{\Psi}_{-} \leftarrow HDMD\left(\tilde{{\bf Y}}\right)$\;
Using $\bar{{\bf K}}_{M},~{\bf K}_{a}, ~\mathbf{\Psi}_{-}$, compute $\mathcal{L}_{pred}$ and $\mathcal{L}_{dmd}$\;
Compute $\mathcal{D}(\tilde{{\bf Y}})$ and then find $\mathcal{L}_{recon}$\;
$\mathcal{L}_{tot} \leftarrow \mathcal{L}_{recon} + \mathcal{L}_{pred} + \mathcal{L}_{dmd} + \alpha \mathcal{L}_{reg}$\;
Using optimizer, attempt to reduce $\mathcal{L}_{tot}$\;
}
{\bf Update for $\bar{N}_{ob}$ }: All quantities computed for fixed network weights and subset of training data.\\
Compute (as above) $\mathcal{L}_{tot}(\bar{N}_{ob})$\;
$\mathcal{L}_{min} \leftarrow \mathcal{L}_{tot}(\bar{N}_{ob})$\;
\For{$n \leftarrow \bar{N}_{ob}-1$ \KwTo $\bar{N}_{ob}+1 $}{
Compute $\mathcal{L}_{tot}(n)$\;
\If{$\mathcal{L}_{tot}(n) < \mathcal{L}_{min}$ and $|1-\mathcal{L}_{tot}(n)/\mathcal{L}_{min}|>f_{r}$}{
$\mathcal{L}_{min}\leftarrow \mathcal{L}_{tot}(n)$\;
$\bar{N}_{ob} \leftarrow n$\;
}
}
}
\caption{The DLHDMD Algorithm}
\label{dlhdmd}
\end{algorithm}

\subsection{Results for DLHDMD}
We now show how the DLHDMD performs on several dynamical systems.  We take as our training set $N_{C}=7000$ randomly chosen initial conditions with their affiliated trajectories. For the lower dimensional systems, we use 1000 randomly chosen initial conditions for testing purposes.  For the higher dimensional systems, we use 500 randomly chosen trajectories for testing.  The networks $\mathcal{E}$ and $\mathcal{D}$ consist of 5 layers each.  Two of the layers are for input and output, while the remaining three are hidden nonlinear layers using the ReLU activation function.  Alternate architectures with more or less hidden layers were studied with the present choice being found to lead to the most efficient training.  The training is done in PyTorch \cite{NEURIPS2019_9015} over $E_{max}=200$ epochs for the lower dimensional systems and $E_{max}=100$ epochs for the KS system.  In all cases, the training used an ADAM optimizer with learning rate $\gamma$.   

\subsubsection{Hyperparameter Tuning}
Before runtime then, the user must specify $N_{e}$, $N_{st}$, $N_{B}$, $\alpha$, $\gamma$, $f_{r}$, and an initial choice for $\bar{N}_{ob}$. $N_{st}$ was chosen relative to considerations of the underlying dynamical systems themselves as explained later.  For our tuning results, we found that setting $N_{e}=N_{s}$ and letting our adaptive scheme update $\bar{N}_{ob}$ was generally sufficient to produce useful suggestions for parameter choices.  This also helped manage the computational cost of tuning.  To determine optimal choices of $N_{B}$, $\alpha$, and $\gamma$, we used the Ray tuner \cite{liaw2018tune} set to perform hyperband optimization across six samples at a time.  The batch size $N_{B}$ was chosen from the values $\left\{64, 128, 256\right\}$, $\alpha$ was chosen from a log-uniform sampling of values between $10^{-12}$ and $10^{-8}$, and $\gamma$ was chosen from a log-uniform sampling of values from $10^{-5}$ and $10^{-2}$.  For Lorenz-63 and Rossler, we started with $\bar{N}_{ob}=10$ while for the KS system, we started from $\bar{N}_{ob}=5$.  Significantly smaller or larger choices for the initial value of $\bar{N}_{ob}$ in all cases resulted in less than desirable performance as determined through trial and error.  We also note that when tuning, we chose a fraction of $E_{max}$, say $E_{tst}$, so as to facilitate a more rapid tuning process.  While this did not need to be especially large for the Lorenz or KS systems, Rossler was a different matter, and small choices of $E_{tst}$ lead to very unstable training.  

Determining good choices of $f_{r}$ turned out to be the most difficult task.  Generally speaking, after trial and error, we found that one wanted $f_{r}$ large enough so that the model did not spuriously increase $\bar{N}_{ob}$ in place of forcing the encoder and decoder to better learn the dynamics.  Thankfully, we consistently observed that if $\bar{N}_{ob}=N_{st}$, which corresponds to the models losing any predictive capacity, during training, then $\mathcal{L}_{dmd}$ would drop several orders of magnitude, thereby becoming essentially irrelevant.  Thus if we observe during training that $\bar{N}_{ob}\geq N_{st}$, this clearly indicates that we should increase $f_{r}$.  That said, we also need to keep $f_{r}$ small enough so that the model can take advantage of increasing $\bar{N}_{ob}$ so as to stabilize training and improve model performance.  A strategy to address this issue was to auto-tune for relatively small values of $E_{tst}$ for low choices of $f_{r}$ which then let us get a reasonable estimate on good choices for $\gamma$, $\alpha$, and $N_{B}$.  We would then run these models and see if our choice of $f_{r}$ allowed $\bar{N}_{ob}=N_{st}$ during longer training runs.  If so, we would raise the value of $f_{r}$ and rerun our model, and if this produced a model with low and converging $\mathcal{L}_{tot}$ while keeping $\bar{N}_{ob}<N_{st}$, we kept those parameter choices.  Otherwise, we would retune across another six samples at a higher value of $f_{r}$ and possibly $E_{tst}$.  For each system, this amounted to about three to four tuning experiments overall to produce a good results.  This approach did allow us to ultimately converge on parameter choices which produced excellent results, but we also acknowledge that this is a time consuming process.  Clearly, developing a loss function which helps manage choosing good $f_{r}$ values is desirable and will be a subject of future research.  

The results for all of our experiments are presented in Table \ref{tab:hypterms}.  As can be seen, the underlying differences in dynamics manifest as significant changes to the batch size and learning rate.  Most notably, Rossler's multiscale features demand both a longer number of testing epochs, larger batch size, and lower learning rate reflecting the greater difficulty that the machine has in learning a reasonable optimization path.  From this, as a general rule of thumb, when all else fails, raising $N_{B}$ and lowering $\gamma$ gives the optimization routine more chance to find training paths.  
\begin{table}
\centering
\begin{tabular}{c|c|c|c}
 & Lorenz-63 & Rossler & KS \\
\hline
$\bar{N}_{ob}$ (te) & 10 & 10 & 5\\
$f_{r}$ (te) & .05 & .25 & .15\\
$E_{tst}$ (te) & 20 & 40 & 10\\ 
$N_{B}$ (tn) & 64 & 256 &  64\\
$\alpha$ (tn) & $3.46\times 10^{-12}$ & $2.52\times 10^{-12}$ & $4.85 \times 10^{-12}$\\
$\gamma$ (tn) &  .000507  & .0001 & .00081
\end{tabular}
\caption{Hyperparamter choices and tuning results for the several systems studied.  (te) denotes a parameter chosen through trial-and-error while (tn) denotes one determined via tuning relative to the trial-and-error choices made.  We emphasize that $E_{tst}$ is not used in the final learning algorithm, but it is necessary to chose in order to get useful tuning results.}
\label{tab:hypterms}
\end{table}

Finally, we briefly address the performance of our algorithm with respect to the system size $N_{s}$.  For Lorenz-63 and Rossler, $N_{s}=3$, while for the KS system, $N_{s}=12$.  This is a significant difference, but it does raise the question of whether our method has an upper limit beyond which it can no longer readily train.  To find this, we studied the Lorenz-96 equations \cite{lorenz_2006} for $N_{s}=5$, $10$, and $20$.  We were able to get successful training for all but $N_{s}=20$. Even by following our previous advice of increasing $N_{B}$ and decreasing $\gamma$, we could not find a combination of parameters which allowed for reasonable training.  Why this is the case, and possible remedies, will be a subject of future research. 

\subsubsection{DLHDMD for the Lorenz-63 System}
\begin{figure}[!h]
\centering
\includegraphics[width=5in, height=2in]{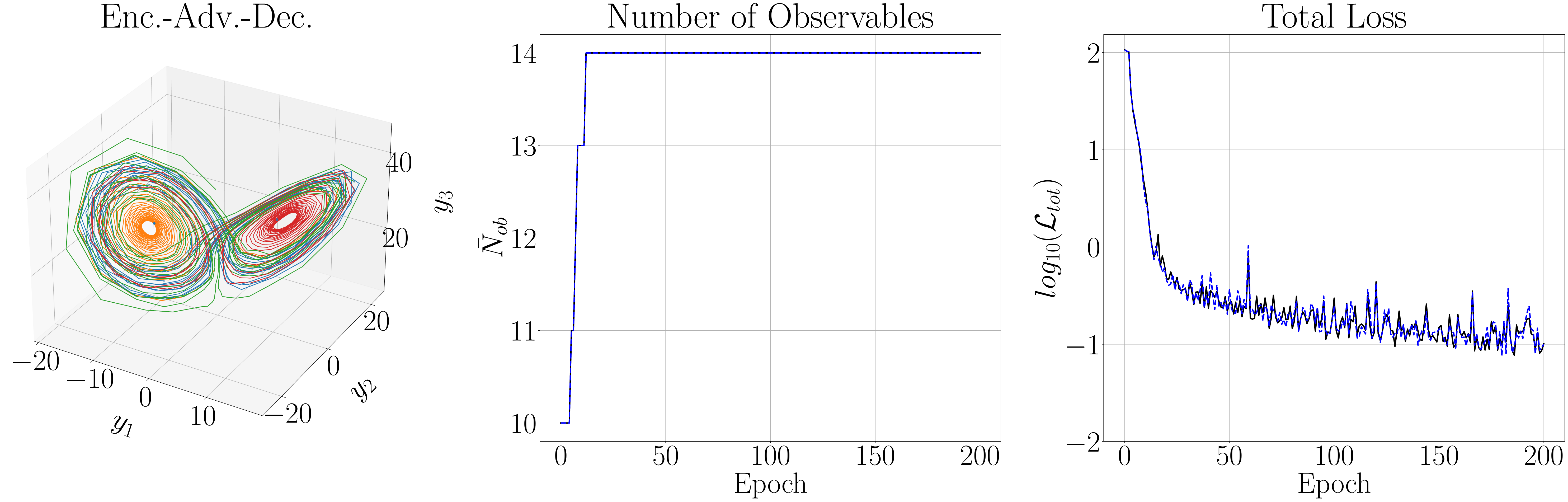}
\caption{Results of the DLHDMD on the Lorenz-63 system after 200 epochs of training for $N_{e}=3$ and $f_{r}=.05$.  From left to right, we see the reconstruction generated by DLHDMD, i.e. $\mathcal{D} \left( \bar{{\bf K}}_{M}{\bf K}_{a}^{N_{st}}\mathbf{\Psi}_{-}\right)$, and the plots of $\bar{N}_{ob}$ and $\mathcal{L}_{tot}$ over epochs.   The reconstruction is generated for times $1\leq t\leq t_{f,w}$ and forecasting is done for times $t_{f,w}\leq t \leq t_{f,w}+1$.  Error plots are over both training (blue/dash) and testing (black/solid) data.  The reconstruction in the leftmost figure is generated over testing data.}
\label{fig:lorenztots}
\end{figure}

\begin{figure}[!h]
\centering
\includegraphics[width=5in, height=2in]{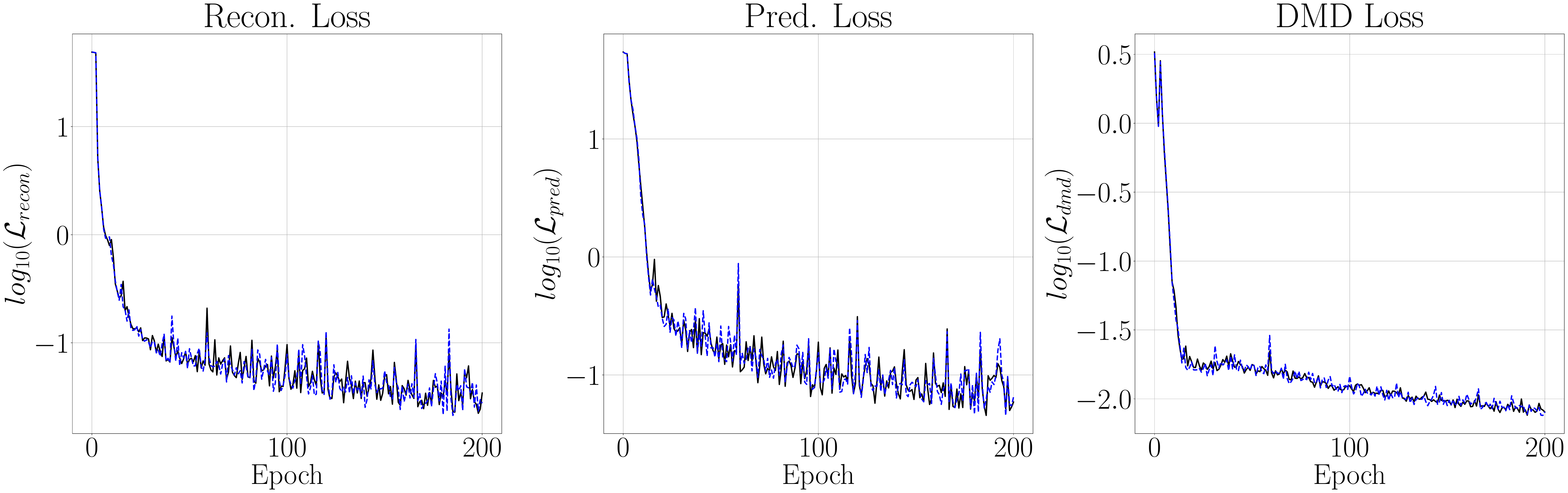}
\caption{Results of the DLHDMD on the Lorenz-63 system after 200 epochs of training for $N_{e}=3$ and $f_{r}=.05$.  Moving from left to right, we plot $\mathcal{L}_{recon}$, $\mathcal{L}_{pred}$, and $\mathcal{L}_{dmd}$.  Error plots are over both training (blue/dash) and testing (black/solid) data.}
\label{fig:lorenzcomps}
\end{figure}

The results of running the DLHDMD for the Lorenz-63 system are found in Figures \ref{fig:lorenztots} and \ref{fig:lorenzcomps}.  The maximum positive Lyupanov exponent, say $\lambda_{L}$, for the Lorenz-63 system can be numerically computed using the methods of \cite{abarbanel}, and we find that $\lambda_{L}\approx .8875$.  In this case then, our prediction window is only slightly less than $1/\lambda_{L}\approx 1.127$, so that we are making predictions up to the point where the strange attractor would tend to induce significant separations in what were initially nearby trajectories.  First taking $N_{e}=3$ and $f_{r}=.05$, as can be seen in Figure \ref{fig:lorenztots}, the overall reconstruction and forecast, plotted for times $t$ such that $1\leq t\leq t_{f,w}+1$, shows excellent agreement with the plot of the Lorenz Butterfly in the top row of Figure \ref{fig:lorenzapprox}.  This degree of accuracy is quantified by the graph of $\mathcal{L}_{pred}$ in Figure \ref{fig:lorenzcomps}, which shows a relative accuracy of about $.1\%$ by the $200^{th}$ epoch.  To achieve this, we see that the DLHDMD progressively raises the value of $\bar{N}_{ob}$.  As seen in Figure \ref{fig:lorenztots}, this process continues until about the $25^{th}$ epoch, at which point $\bar{N}_{ob}=14$ and then stays at this value for the remaining epochs.  What is particularly striking is that $\bar{N}_{ob}<N_{st}$ for the length of training, which means that our trained model is able to make novel forecasts while still providing excellent reconstructions.  We note that larger choices of $f_{r}$ were examined, and generally they prevented the model from updating $\bar{N}_{ob}$ and thus degrading model performance.  

\begin{figure}[!h]
\centering
\includegraphics[width=5in, height=2in]{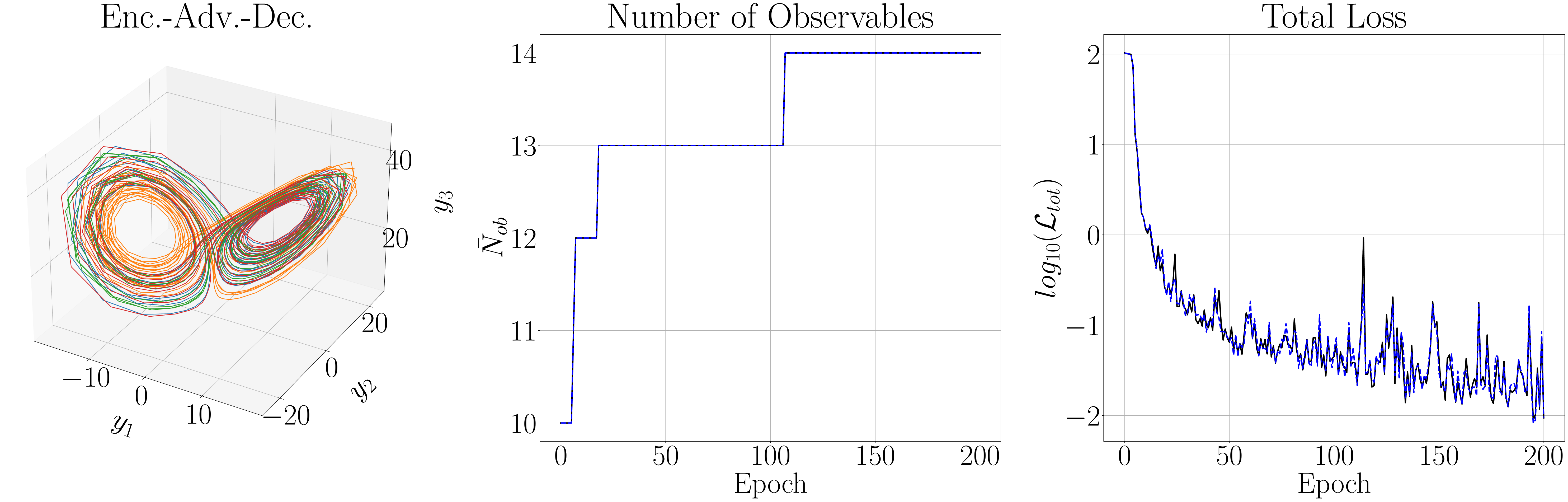}
\caption{Results of the DLHDMD on the Lorenz-63 system after 200 epochs of training for $N_{e}=6$ and $f_{r}=.1$.  From left to right, we see the reconstruction generated by DLHDMD, i.e. $\mathcal{D} \left( \bar{{\bf K}}_{M}{\bf K}_{a}^{N_{st}}\mathbf{\Psi}_{-}\right)$, and the plots of $\bar{N}_{ob}$ and $\mathcal{L}_{tot}$ over epochs.   The reconstruction is generated for times $1\leq t\leq t_{f,w}$ and forecasting is done for times $t_{f,w}\leq t \leq t_{f,w}+1$.  Error plots are over both training (blue/dash) and testing (black/solid) data.  The reconstruction in the leftmost figure is generated over testing data.}
\label{fig:lorenztotsne6}
\end{figure}

\begin{figure}[!h]
\centering
\includegraphics[width=5in, height=2in]{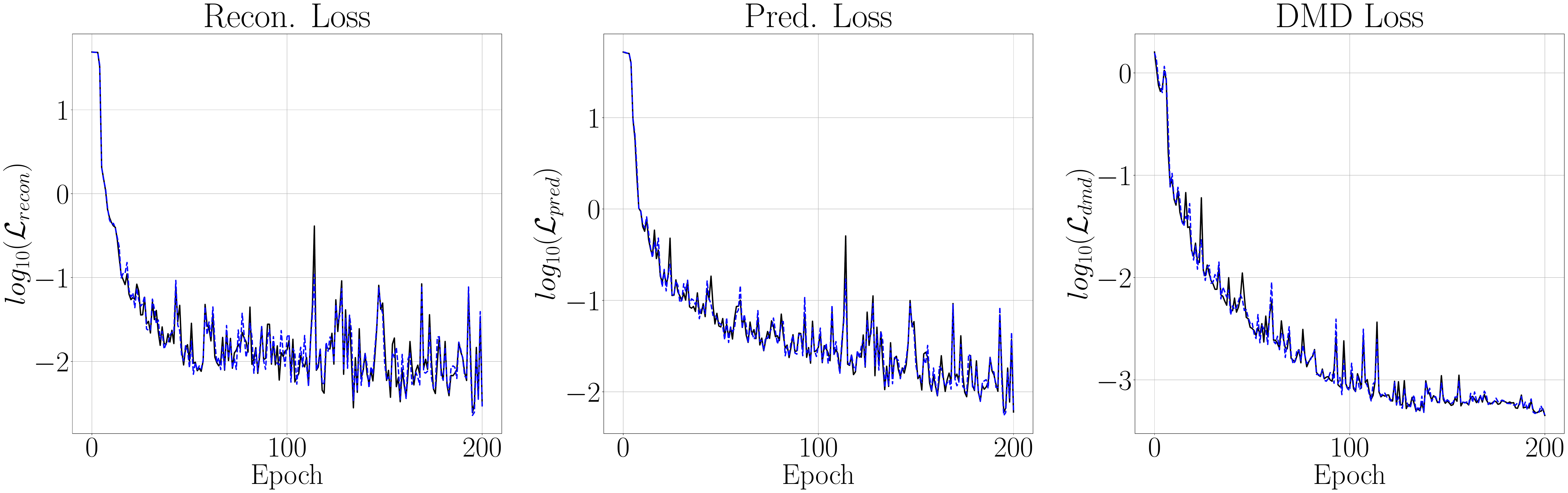}
\caption{Results of the DLHDMD on the Lorenz-63 system after 200 epochs of training for $N_{e}=6$ and $f_{r}=.1$.  Moving from left to right, we plot $\mathcal{L}_{recon}$, $\mathcal{L}_{pred}$, and $\mathcal{L}_{dmd}$.  Error plots are over both training (blue/dash) and testing (black/solid) data.}
\label{fig:lorenzcompsne6}
\end{figure}

We now examine the effect of increasing the embedding dimension so that $N_{e}=6$.  We first note that we had to increase $f_{r}=.1$ in order to keep $\bar{N}_{ob}$ from reaching and surpassing $N_{st}$, which it rapidly did when $f_{r}=.05$.  Otherwise we keep all of the other parameters the same for the sake of comparison.  That said, as can be seen in Figures \ref{fig:lorenztotsne6} and \ref{fig:lorenzcompsne6}, while the training exhibits larger variance around the net trend in the loss, the model is able to converge to an even lower overall loss while still only needing $\bar{N}_{ob}=14$ observables.  However, the difference is not even an order of magnitude better, and this generally reflects that the performance of our models is not drastically affected by changing the embedding dimension.  We also note that the average training time per epoch when $N_{e}=3$ is 22 s while for $N_{e}=6$ it is 44 s.  We then see that raising the embedding dimension is a fraught affair.  While it can improve model performance, it becomes much more computationally costly for what appear to be relatively nominal returns.  

\begin{figure}[!h]
\centering
\includegraphics[width=4in, height=2.in]{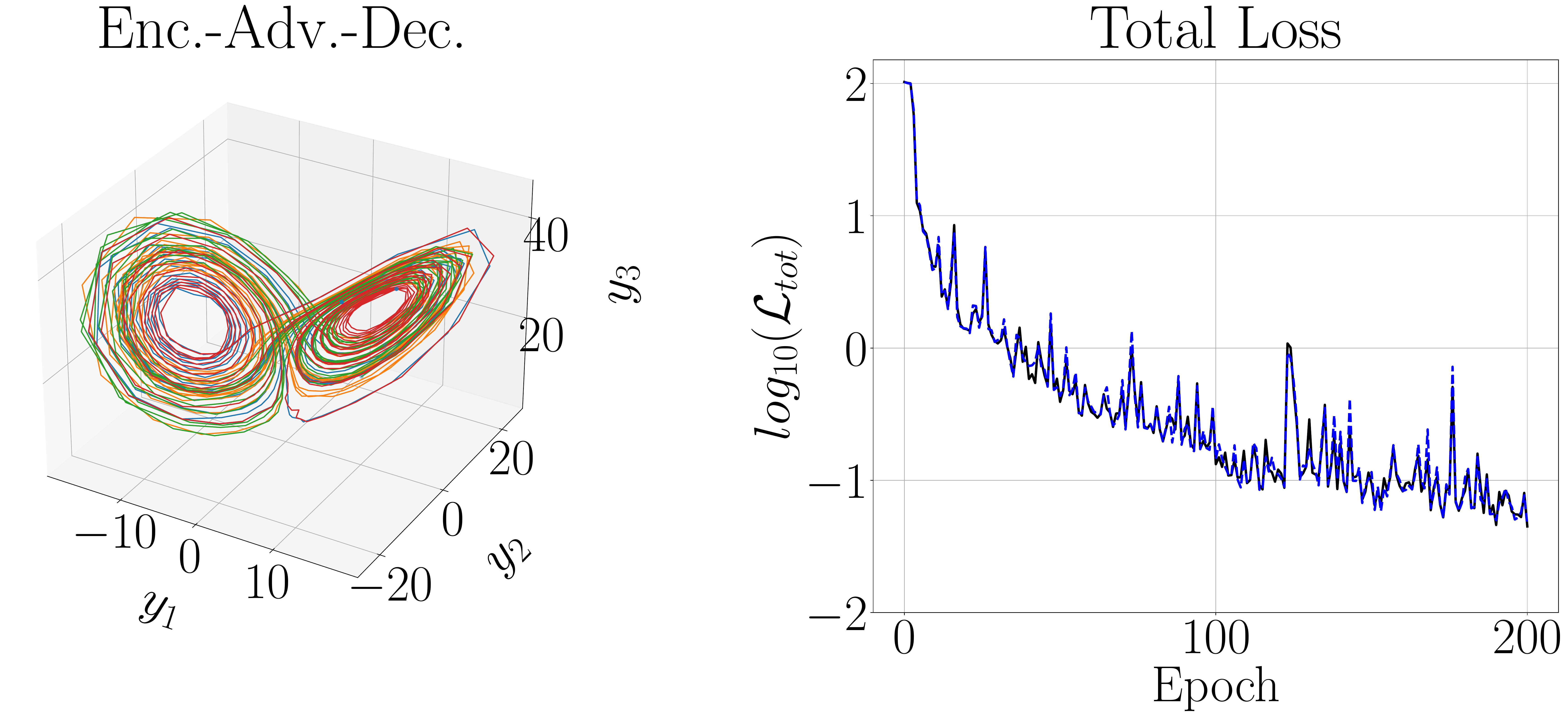}\\
\caption{Results of the DLHDMD on the Lorenz-63 system after 200 epochs of training without updating the initial choice of $\bar{N}_{ob}=10$ for $N_{e}=6$ and $f_{r}=\infty$.  From left to right, we see the reconstruction generated by DLHDMD, i.e. $\mathcal{D} \left( \bar{{\bf K}}_{M}{\bf K}_{a}^{N_{st}}\mathbf{\Psi}_{-}\right)$, and the plot of $\mathcal{L}_{tot}$ over epochs.   The reconstruction is generated for times $1\leq t\leq t_{f,w}$ and forecasting is done for times $t_{f,w}\leq t \leq t_{f,w}+1$, where $t_{f,w}=19.5$.  Error plots are over both training (blue/dash) and testing (black/solid) data.  The reconstruction in the leftmost figure is generated over testing data.}
\label{fig:gdldmd_nesix}
\end{figure}

Of course, one might wonder to what extent increasing $N_{e}$ can be exchanged for increasing $\bar{N}_{ob}$.  Thus, we set $f_{r}=\infty$ so that we do not perform any update of $\bar{N}_{ob}$ during training.  The results of doing this for $N_{e}=6$ can be seen in Figure \ref{fig:gdldmd_nesix}.  Comparing $\mathcal{L}_{tot}$ across Figures \ref{fig:lorenztots}, \ref{fig:lorenztotsne6}, and \ref{fig:gdldmd_nesix}, we see that increasing $N_{e}$ but not updating $\bar{N}_{ob}$ gets us a model of intermediate accuracy, though again the overall differences are relatively nominal relative to order of magnitudes present in the original data sets.  Ultimately then, we see that the updating scheme controlled by $f_{r}$ improves model performance, though the update threshold needs to be chosen carefully to be large enough to make sure changing $\bar{N}_{ob}$ does not preclude the encoder/decoder networks from doing the harder work of learning the dynamics while at the same time not choosing $f_{r}$ so large that no update moves are made.  Finally, we refer the reader to the Appendix where we examine not only setting $f_{r}=\infty$, but also fixing $\bar{N}_{ob}=1$, thereby ``turning off" Hankel DMD in our algorithm.  We also compare using the global EDMD method with a local one, which then essentially reduces to the DLDMD method of \cite{lago_dldmd}.  In all cases, we find the results significantly worse than what is presented in this section, thereby showing our additions to the DLDMD method have markedly improved model development.  

\subsubsection{DLHDMD for the Rossler System}
We now study the Rossler system given by 
\begin{align*}
\dot{y}_{1} = & -y_{2}-y_{3},\nonumber \\ 
\dot{y}_{2} = & y_{1} + ay_{2},\nonumber \\ 
\dot{y}_{3} = & b + y_{3}\left(y_{1}-c\right),  \label{lorenzequation}
\end{align*}
where
\[
a =  .1, ~ b = .1, ~ c = 14.
\]
Aside from the dynamics coalescing onto a strange attractor, the disparity in parameter values gives rise to multiscale phenomena so that there are slow and fast regimes of the dynamics.  The slow portions are approximated by harmonic motion in the $(y_{1},y_{2})$ plane, and the fast portion moves along the $y_{3}$ coordinate.  This strong disparity in time scales also appears by way of $\lambda_{L}\approx 1.989$, which is more than double the maximal Lyupanov exponent for the Lorenz-63 system.  Thus dynamics separate along the strange attractor twice as fast.  

\begin{figure}[!h]
\centering
\includegraphics[width=5in, height=2in]{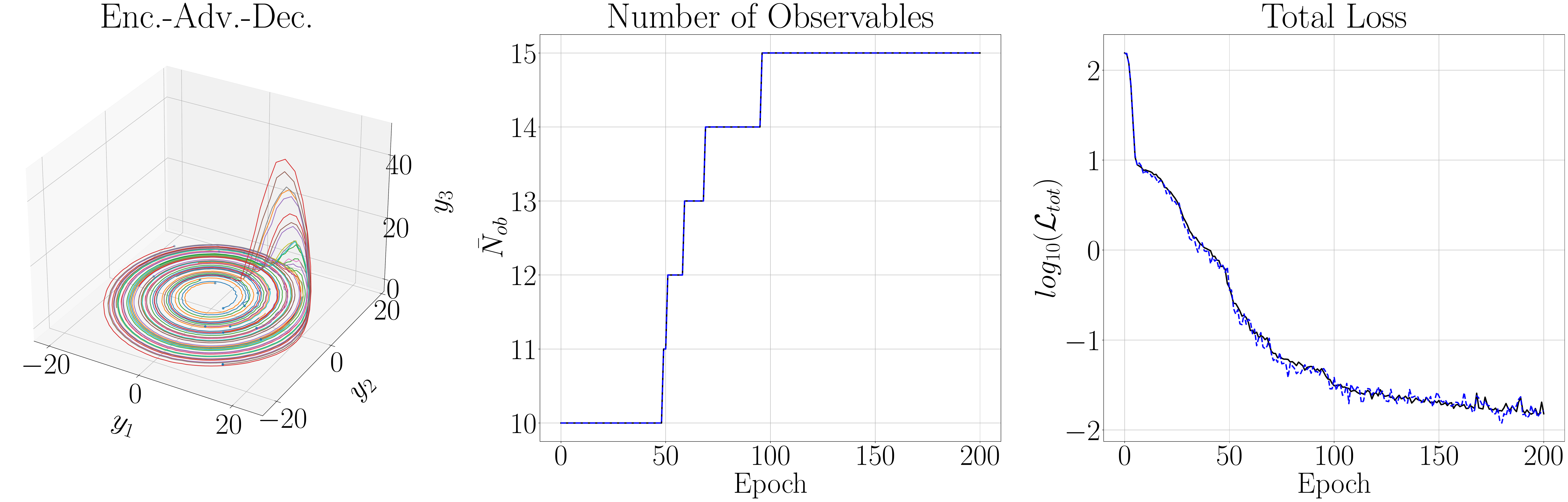}
\caption{Results of the DLHDMD on the Rossler system after 200 epochs of training for $N_{e}=3$ and $f_{r}=.25$.  From left to right, we see the reconstruction generated by DLHDMD, i.e. $\mathcal{D} \left( \bar{{\bf K}}_{M}{\bf K}_{a}^{N_{st}}\mathbf{\Psi}_{-}\right)$, and the plots of $\bar{N}_{ob}$ and $\mathcal{L}_{tot}$ over epochs.   The reconstruction is generated for times $1\leq t\leq t_{f,w}$ and forecasting is done for times $t_{f,w}\leq t \leq t_{f,w}+1$.  Error plots are over both training (blue/dash) and testing (black/solid) data.  The reconstruction in the leftmost figure is generated over testing data.}
\label{fig:rosslertots}
\end{figure}

\begin{figure}[!h]
\centering
\includegraphics[width=5in, height=2in]{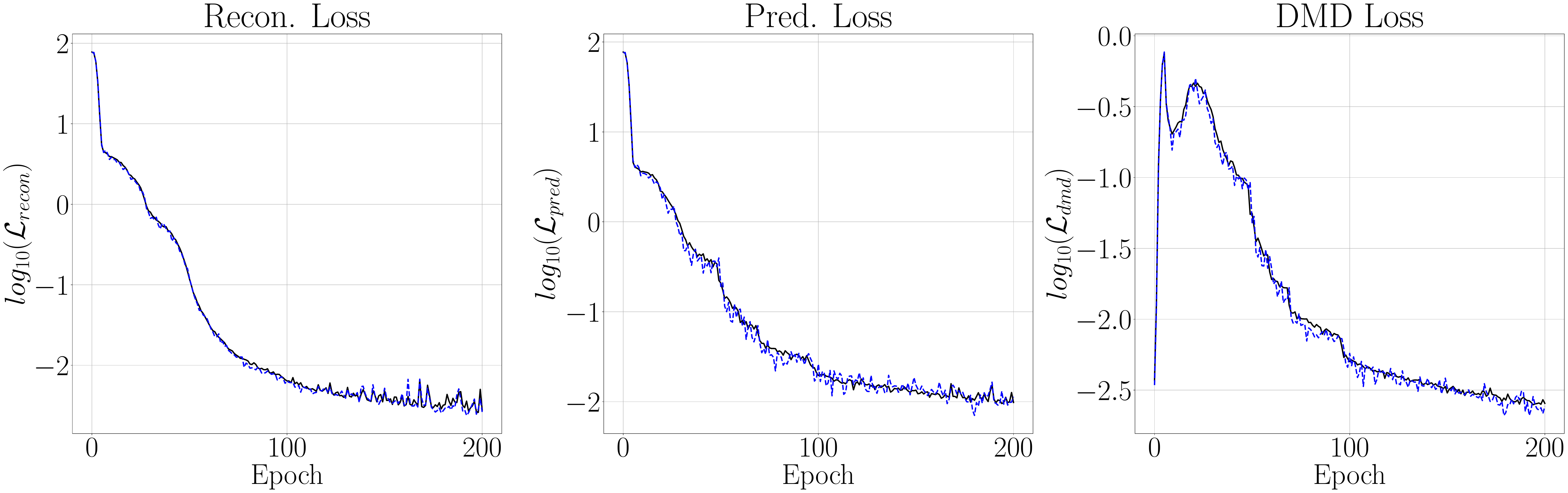}
\caption{Results of the DLHDMD on the Rossler system after 200 epochs of training for $N_{e}=3$ and $f_{r}=.25$.  Moving from left to right, we plot $\mathcal{L}_{recon}$, $\mathcal{L}_{pred}$, and $\mathcal{L}_{dmd}$.  Error plots are over both training (blue/dash) and testing (black/solid) data.}
\label{fig:rosslercomps}
\end{figure}

Using the parameter choices in Table \ref{tab:hypterms} and taking $N_{e}=3$, in Figures  \ref{fig:rosslertots} and \ref{fig:rosslercomps} we get the training and testing results of DLHDMD.  We immediately observe that, relative to the Lorenz-63 system, the much larger choice of $N_{B}$ and much smaller choice of learning rate $\gamma$ produce a far less noisy training/validation curve; see in particular Figure \ref{fig:rosslercomps}.  Somewhat suprisingly, we are able to train to overall lower loss values even with the significantly larger choice for $f_{r}$.  We note that even nominally smaller choices of $f_{r}$ resulted in the collapse of the $\mathcal{L}_{dmd}$ term, so the slow/fast dynamics create very concrete differences between the Rossler and Lorenz-63 models.  That said, we are still able to keep the final value of $\bar{N}_{ob}<N_{st}$ and still get excellent reconstruction, so it is clear that our overall approach is working very well.  We also note that we examined the impact of letting $N_{e}=6$ and generally found no significant improvement in training or testing.  In this case then, we see that there is no hard and fast rule about using higher embedding dimensions to improve model performance, and so it must be treated as a general hyperparameter which can induce significant computational cost.    

\subsubsection{DLHDMD for the KS Equation}
To see the edges of our method, we now examine spatio-temporal chaos generated by the KS equation with periodic-boundary conditions in the form
\[
u_{t} + u_{xx} + u_{xxxx} + uu_{x} = 0, ~ u(x+2L,t) = u(x,t).
\]
Note, given the vast size of the literature around the KS equation, we refer the reader to \cite{robinson} for an extensive bibliography with regards to details and pertinent proofs of facts used in this section.  Introducing the rescalings
\[
\tilde{t} =\frac{t}{T}, ~ \tilde{x} = \frac{\pi}{L}x, ~ u = A\tilde{u},
\]
and taking the balances
\[
A = \frac{L}{\pi T}, ~ T = \left(\frac{L}{\pi}\right)^{2}, 
\]
we get the equivalent KS equation (dropping tildes for ease of reading)
\[
u_{t} + u_{xx} + \nu u_{xxxx} + uu_{x} = 0, ~ \nu = \left(\frac{\pi}{L}\right)^{2}.
\]
Looking at the linearized dispersion relationship $\omega(k) = k^{2} - \nu k^{4}$, we see that the $\nu$ parameter acts as a viscous damping term.  Thus, as the system size $L$ is increased, the effective viscosity is decreased, thereby allowing for more complex dynamics to emerge.  As is now well known, for $L$ sufficiently large, a fractional-dimensional-strange attractor forms which both produces intricate spatio-temporal dynamics while also allowing for a far simpler representation of said dynamics.  It is has been shown in many different works (see for example \cite{citanovic}) that $L=11$ generates a strange attractor with dimension between eight and nine, and that this is about the smallest value of $L$ which is guaranteed to generate chaotic dynamics.  We therefore set $L=11$ throughout the remainder of this section.  

To study the DLDHMD on the KS equation, we use KS data numerically generated by a pseudo-spectral in space and fourth-order exponential-differencing Runge-Kutta in time method \cite{kassam} of lines approach.  For the pseudo-spectral method, $K=128$ total modes are used giving an effective spatial mesh width of $2L/K = .172$, while the time step for the Runge-Kutta scheme is set to $\delta t = .25$.  These particular choices were made with regards to practical memory and simulation time length constraints.  After a burn in time of $t_{b}=\left(L/\pi\right)^{4} = 150.3$, which is the time scale affiliated with the fourth-order spatial derivative for the chosen value of $L$, 8000 trajectories of total simulation time length $t_{f} = \left(L/\pi\right)^{4}$ were used with gaps of $L/\pi$ in between to allow for nonlinear effects to make each sample significantly different from its neighbors.  Each of the 8000 space/time trajectories was then separated via a POD into space and time modes; see \cite{berkooz}.  Taking $N_{s}=12$ modes captured between 97.8\% and 99.4\% of the total energy.  The choice of the total time scale $t_{f}$ also ensured that the ratio of the largest and smallest singular values affiliated with the POD was between $10^{-1.1}$ and $10^{-1.9}$ so that the relative importance of each of the modes was roughly the same across all samples.  We take this as an indication that each 12-dimensional affiliated time series is accurately tracing dynamics along a common finite-dimensional strange attractor as expected in the KS equation.  Again, using the methods of \cite{abarbanel}, we can find across batches that typically the largest positive Lyupanov exponent $\lambda_{L}\approx .3930$, so that $1/\lambda_{L}\approx 2.545$ is the time after which we anticipate that the strange attractor fully pulls trajectories apart. 

\begin{figure}[!h]
\centering
\includegraphics[width=5.in, height = 2.in]{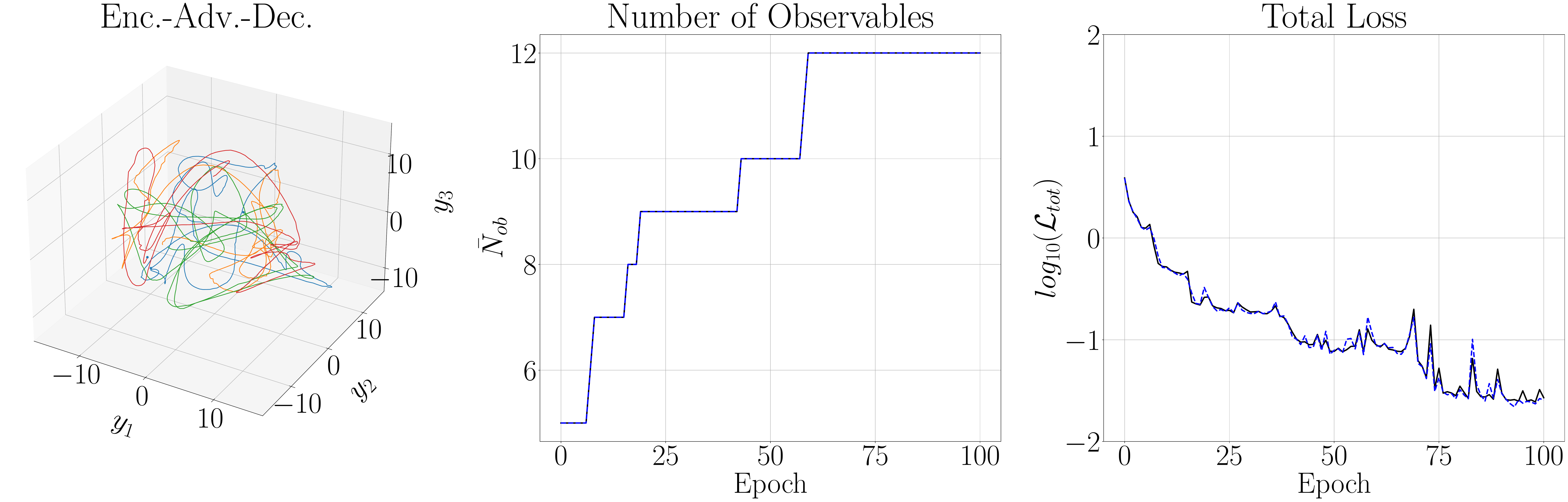}
\caption{Results of the DLHDMD on the KS system after 100 epochs of training.  From left to right, we see the first three dimensions of the reconstruction generated by DLHDMD, i.e. $\mathcal{D} \left( \bar{{\bf K}}_{M}{\bf K}_{a}^{N_{st}}\mathbf{\Psi}_{-}\right)$, and the plots of $\bar{N}_{ob}$ and $\mathcal{L}_{tot}$ over epochs.  Again, the reconstruction is generated for times $L/\pi \leq t\leq t_{f,w}$ and forecasting is done for times $t_{f,w}\leq t \leq t_{f,w}+L/\pi$.  $t_{f,w}$ is initially $(L\pi)^{4}-1.25$.  Error plots are over both training (blue/dash) and testing (black/solid) data.}
\label{fig:ksres_tots}
\end{figure}

\begin{figure}[!h]
\centering
\includegraphics[width=5.in, height = 2.in]{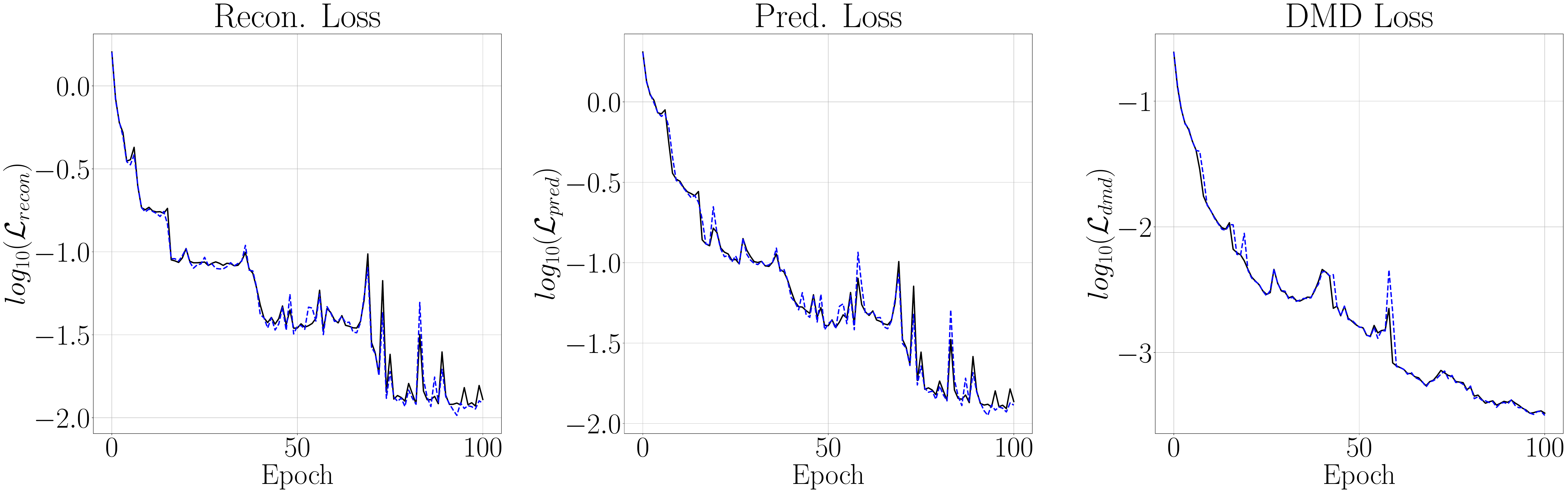}
\caption{Results of the DLHDMD on the KS system after 100 epochs of training.  The reconstruction is generated from $N_{B}=64$ testing trajectories.  Moving from left to right, we plot $\mathcal{L}_{recon}$, $\mathcal{L}_{pred}$, and $\mathcal{L}_{dmd}$.  Error plots are over both training (blue/dash) and testing (black/solid) data.}
\label{fig:ksres_comps}
\end{figure}

\begin{figure}
\centering
\includegraphics[width=1\textwidth]{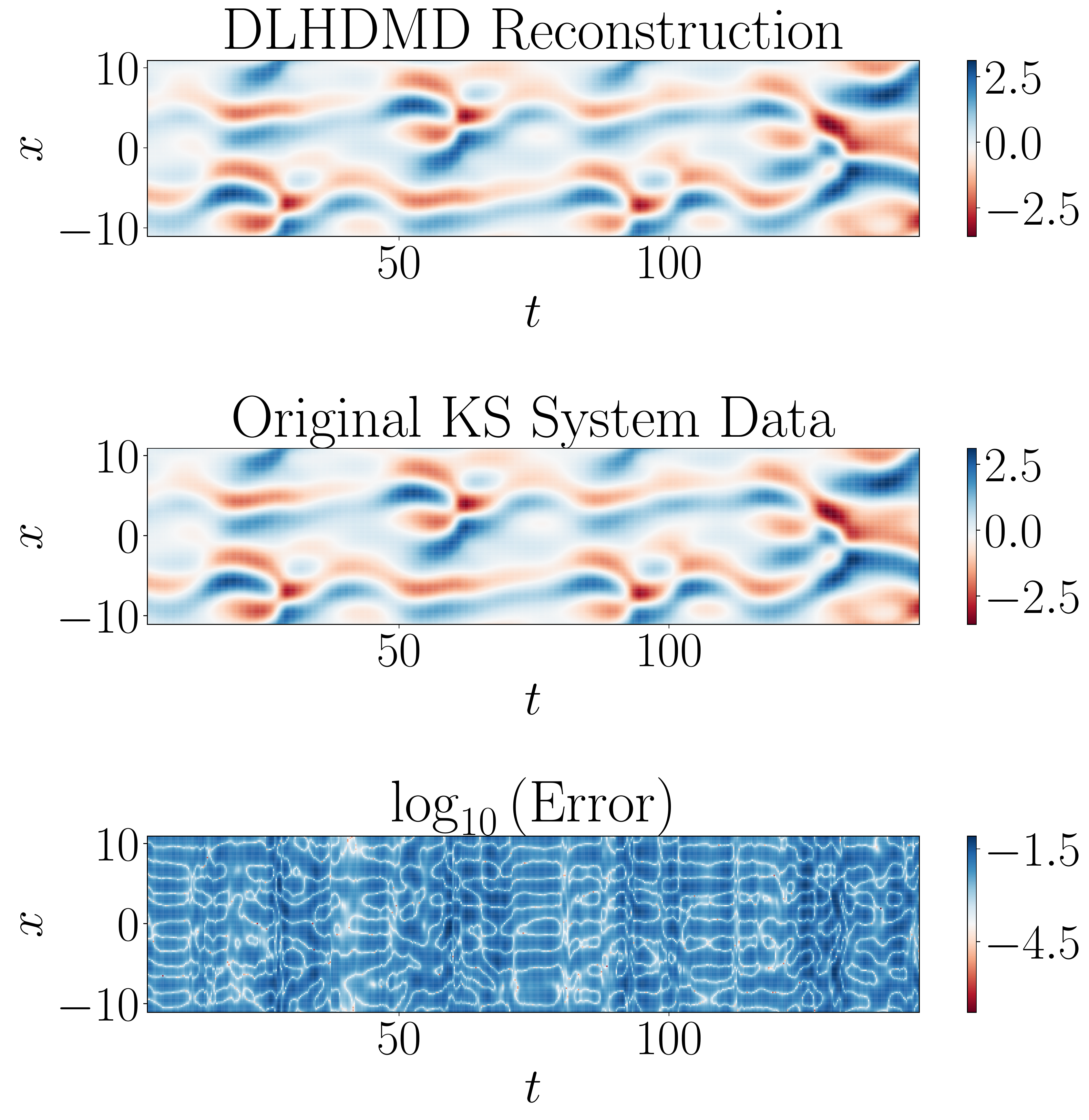}
\caption{Comparison of DLHDMD results and original KS data with the log-scaled error.  The error is overall quite small with peaks in the underlying solution and approximation being the source of any real measurable difference.}
\label{fig:kscomp}
\end{figure}

With regards to the details of the DLHDMD, we again use $N_{C}=7000$ samples for training, and 500 for testing.  The best results with regards to window size were found when we initially set $\bar{N}_{ob}=5$ and $f_{r}=.15$.  The iterated reconstruction/forecast horizon determined by the choice of $N_{st}$ was chosen so that $N_{st}= (L/\pi)/\delta t\approx 14$, corresponding to the time scale over which nonlinear advection acts.  Thus, reconstruction is done on each sample for values of $t$ such that $L/\pi \leq t \leq t_{f,w}$, and iterated reconstruction/forecasting is done for $t$ such that $t_{f,w} \leq t \leq t_{f,w} + L/\pi$.  Note, for our initial choice of $N_{w}$, we have that initially $t_{f,w}=(L\pi)^{4}-1.25$.  The results of DLHDMD training on the $N_{s}=12$ dimensional POD reduction of the KS dynamics is shown in Figures \ref{fig:ksres_tots}, \ref{fig:ksres_comps}, and \ref{fig:kscomp}.  Likewise, our prediction window is longer than the timescale set by $\lambda_{L}$, so we argue our forecasts are over time scales for which chaotic effects are significant.  We see that the reconstruction and predictions appear accurate; see in particular the comparisons in Figure \ref{fig:kscomp}, as seen in Figure \ref{fig:ksres_comps}.  Likewise, by choosing $f_{r}=.15$, we prevent collapse in $\mathcal{L}_{DMD}$ and keep $\bar{N}_{ob}<N_{st}$.  

Overall then, we find that through careful parameter tuning, especially with regards to how one chooses $f_{r}$, we are able to generate models exhibiting high reconstruction accuracy while also providing nontrivial time horizons for prediction of future dynamics, measured by the difference between $\bar{N}_{ob}$ relative to $N_{st}$.  That said, the parameter tuning process can be time consuming, and more automatic methods would be necessary to develop before attempting to improve the results presented in this work.  
\section{Mutual Information for Characterizing Embeddings}
Given the success of the DLHDMD in reconstructing and forecasting dynamics along a strange attractor, especially when compared to the relative failure of trying to do the same using just the HDMD or DLDMD alone, it is of further interest to try to assess exactly what role the auto-encoder plays in improving the outcome of the HDMD.  While we can certainly point to the performance of the components of the loss function $\mathcal{L}_{tot}$ to explain the impact of the encoder, this does not provide us with any more explanatory power.  In \cite{lago_dldmd}, it was empirically shown that the role of the encoder was to generally transform time series to nearly monochromatic periodic signals, which is to say, the effect of encoding was to generate far more localized Fourier spectral representations of the original time series.  This does not turn out to be the case though for the DLHDMD.  Instead, inspired both by the evolving understanding of how mutual information better explains results in dynamical systems \cite{fraser, bollt} and machine learning \cite{tishby1, calin}, we assess the impact of the encoder on the DLHDMD by tracking how the information across dimensions and time lags changes in the original and latent variables.  

For two random variables ${\bf X}$ and ${\bf Y}$ with joint density $p({\bf X},{\bf Y})$, the mutual information (MI) between them $I({\bf X},{\bf Y})$ is defined to be 
\[
I({\bf X},{\bf Y}) = \int p({\bf x},{\bf y})\log\left( \frac{p({\bf x},{\bf y})}{p({\bf x})p({\bf y})}\right)d{\bf x}d{\bf y},
\]
where $p({\bf X})$ and $p({\bf Y})$ are the affiliated marginals.  One can readily show that $I({\bf X},{\bf Y})\geq 0$ and $I({\bf X},{\bf Y}) = 0$ if and only if ${\bf X}$ and ${\bf Y}$ are independent.  Thus information gives us a stronger metric of statistical coupling between random variables than more traditional tools in time series analysis such as correlation measurements.  We also should note here that $I({\bf X},{\bf Y}) = I({\bf Y},{\bf X})$, which is to say it is symmetric.  We also note that MI is invariant under the action of diffeomorphisms of the variables.  Thus we cannot expect to get much use from computing the multidimensional MI of the original and latent variables, thereby allowing for meaningful differences to appear between original and latent variable computations.  

Instead, using the $N_{C}=1000$ trajectories in the test data, we define the $m$-step averaged lagged self-information (ALSI) between the $n^{th}$ and $v^{th}$ dimensions $I_{nv}(m)$ to be 
\[
I_{nv}(m) = \frac{1}{N_{C}}\sum_{k=1}^{N_{C}} I\left(y_{n, \cdot, k},  y_{v, \cdot+m, k}\right).
\]
We refer to the parameter $m$ as a {\it lag}.  In words then, after averaging over the ensemble of initial conditions in the test data, we compute the degree to which the signal becomes statistically independent from itself across all of the dimensions along which the dynamics evolve.  We emphasize that due to the strong nonlinearities in our dynamics, we compute the lagged information as opposed to the more traditional auto-correlation so as to get a more accurate understanding of the degree of self-dependence across dimension in our dynamics.  Further, by measuring the lagged MI across isolated dimensions, we break the invariance of MI with respect to diffeomorphisms.  

\subsection{MI for the Lorenz-63 System}
\begin{figure}[!h]
\centering
\begin{tabular}{cc}
\includegraphics[width=2.25in, height=1.75in]{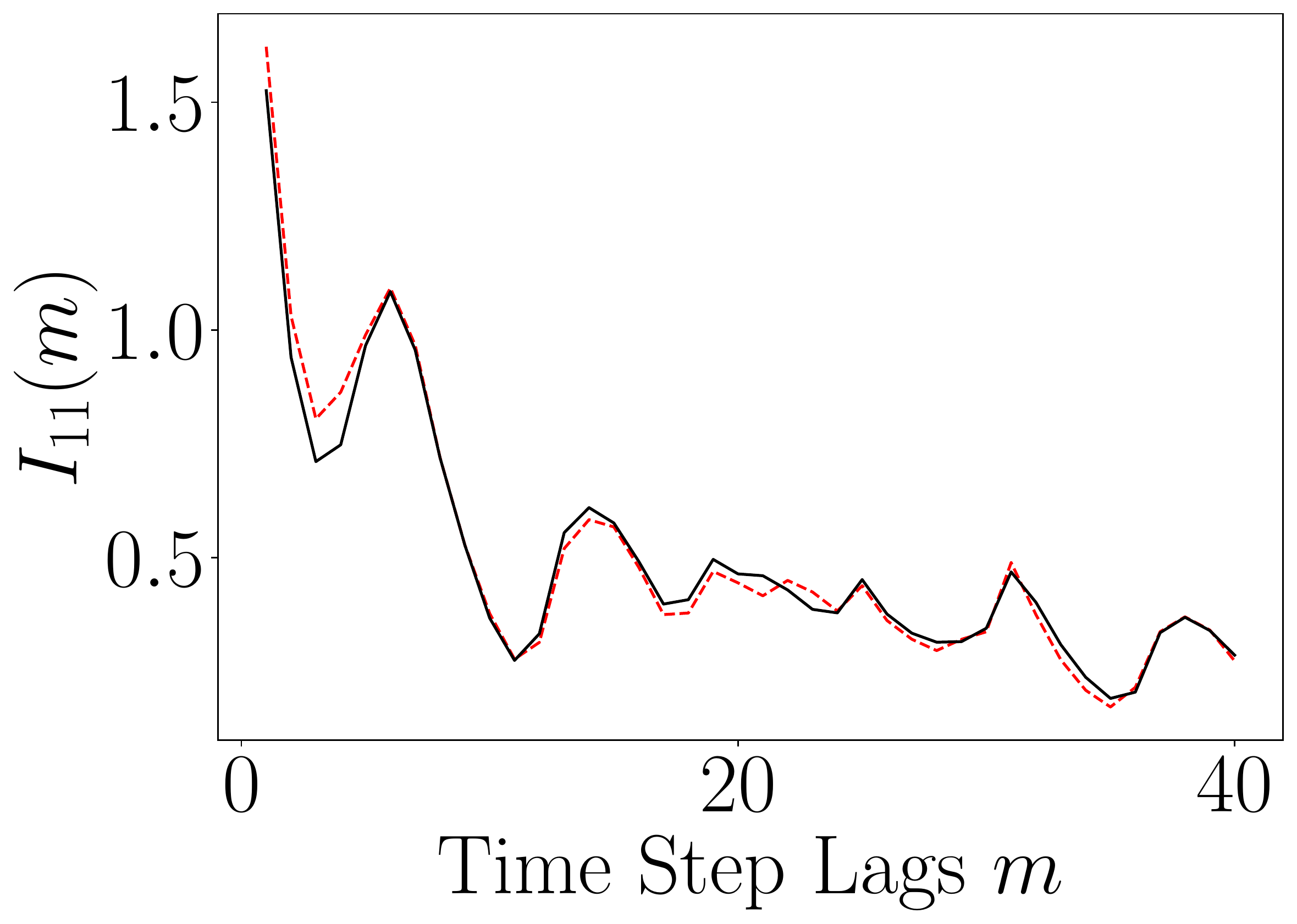} & \includegraphics[width=2.25in, height = 1.75in]{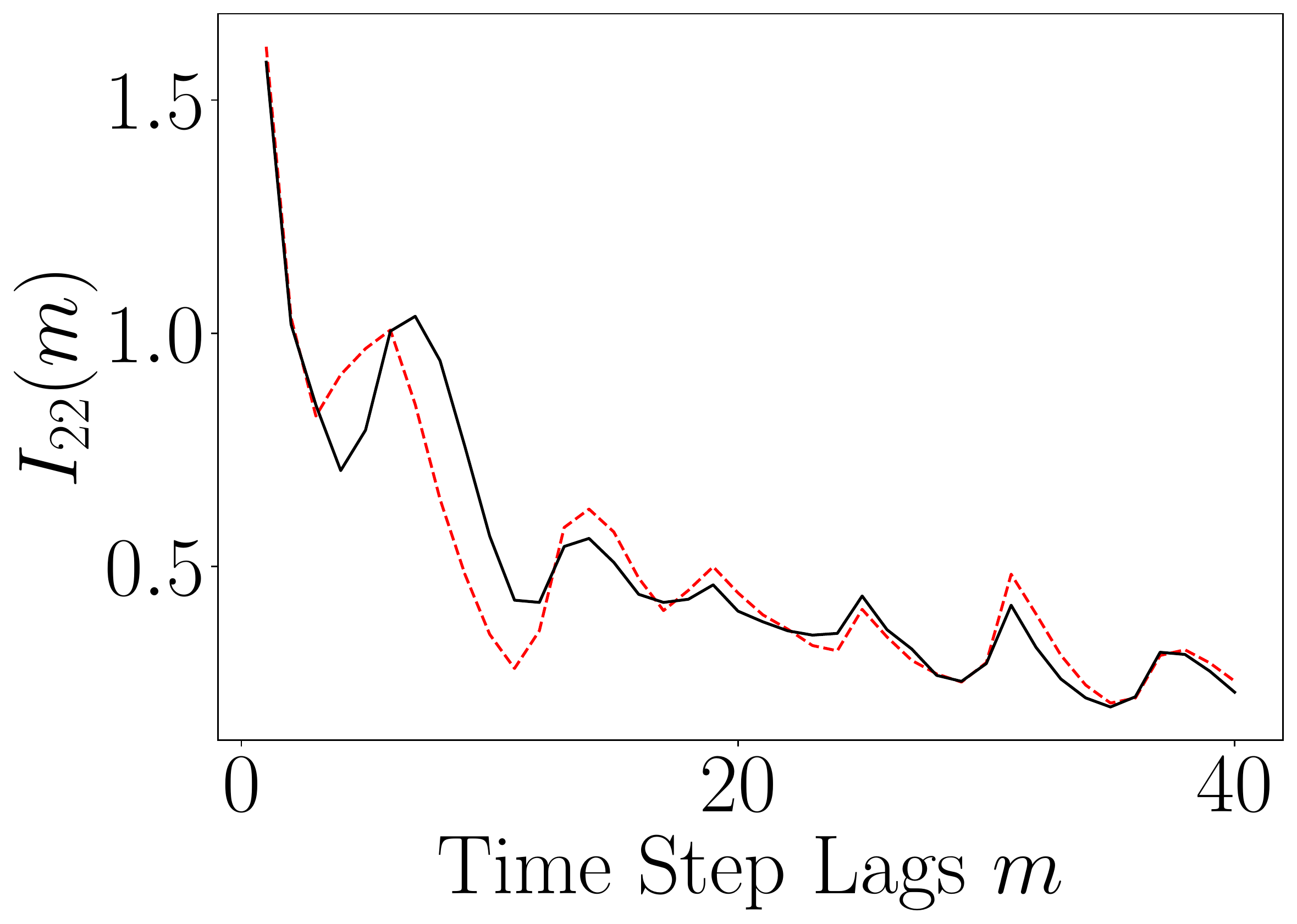} \\
(a)  &  (b) \\
 \includegraphics[width=2.25in, height = 1.75in]{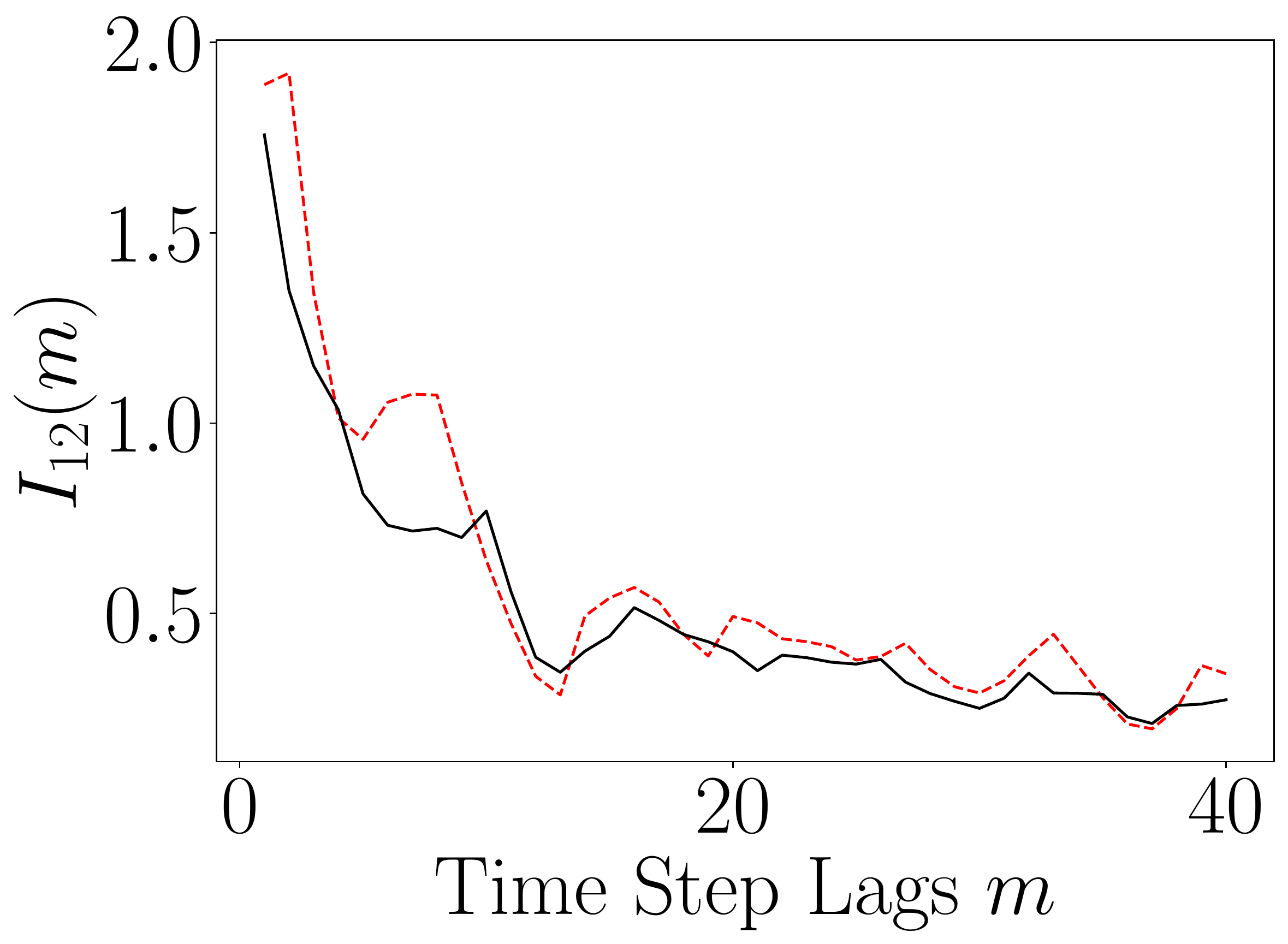} & \includegraphics[width=2.25in, height = 1.75in]{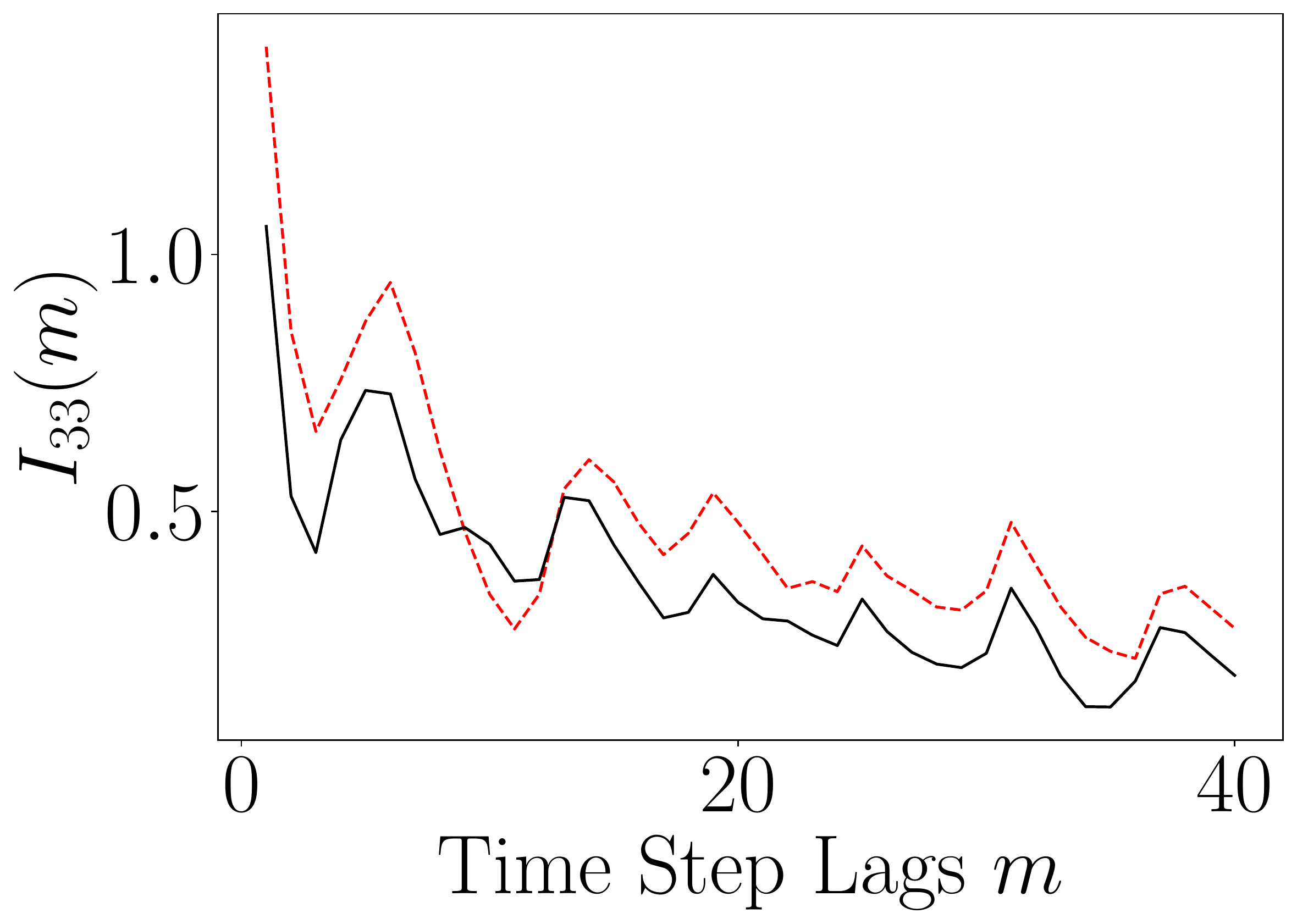} \\
(c) & (d)\\
\includegraphics[width=2.25in, height = 1.75in]{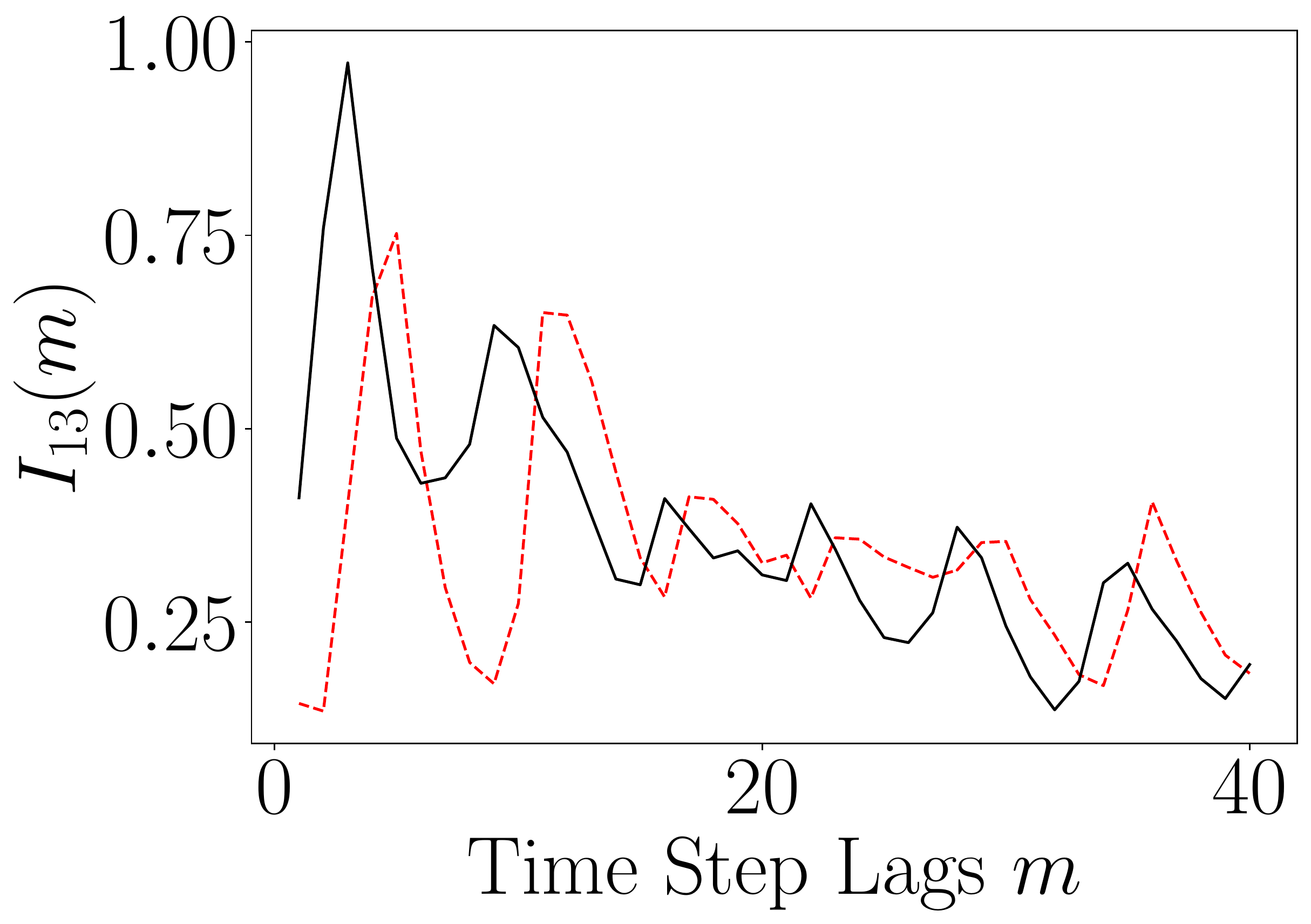} & \includegraphics[width=2.25in, height = 1.75in]{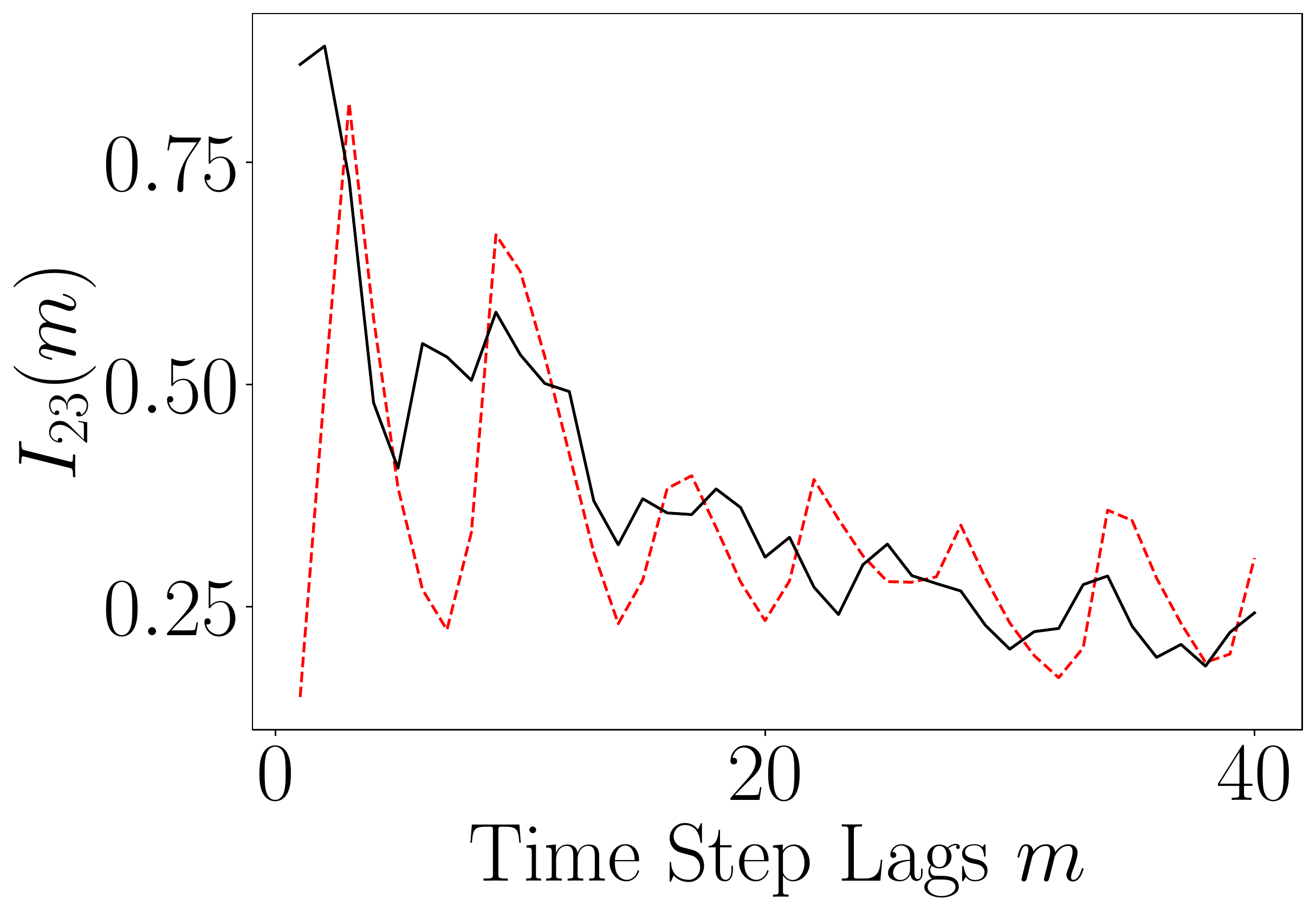}\\
(e) & (f)
\end{tabular}
\caption{For the Lorenz-63 system, plots of the ALSI $I_{nv}(m)$ for $(n,v)=(1,1)$ (a), $(n,v)=(2,2)$ (b), $(n,v)=(1,2)$ (c), $(n,v)=(3,3)$ (d), $(n,v)=(1,3)$ (e), and $(n,v)=(2,3)$ (f) for both the original (red/dash) and latent (black/solid) coordinates.  As can be seen, the encoder tends to reduce the ALSI along each physical dimension aside from those involving the third physical dimension, for which the ALSI is enhanced for shorter lags and decreased for longer ones.}
\label{fig:micomps}
\end{figure}

The results of computing the ALSI for the Lorenz-63 system are plotted in Figure \ref{fig:micomps}.  As can be seen, the impact of the encoder is to either weakly attenuate the dependency between dimensions; see $I_{33}$, or as for $I_{11}$ and $I_{22}$, leave the ALSI essentially unchanged.  Finally, we also see significant phase shifts in the lag count; see $I_{13}$ and $I_{23}$.  In these phase shifts, we see that the shift is always left towards shorter lags, so that the dependence in the latent variables decays more rapidly than in the original variables.  In this sense then, the overall tendency of the encoder is to either reduce MI or cause time series to become more independent more rapidly.  Otherwise though, the timescales of oscillation in the latent variables are essentially identical to those seen in the latent variables.  In terms of the DLHDMD, we might then say that the encoder assists the HDMD by generally making the rows of the affiliated Hankel matrices more independent, especially over longer time scales, and therefore more meaningful with regards to their generating more accurate approximations of the underlying Koopman operator.  

\subsection{MI for the Rossler System}
\begin{figure}[!h]
\centering
\begin{tabular}{cc}
\includegraphics[width=2.25in, height=1.75in]{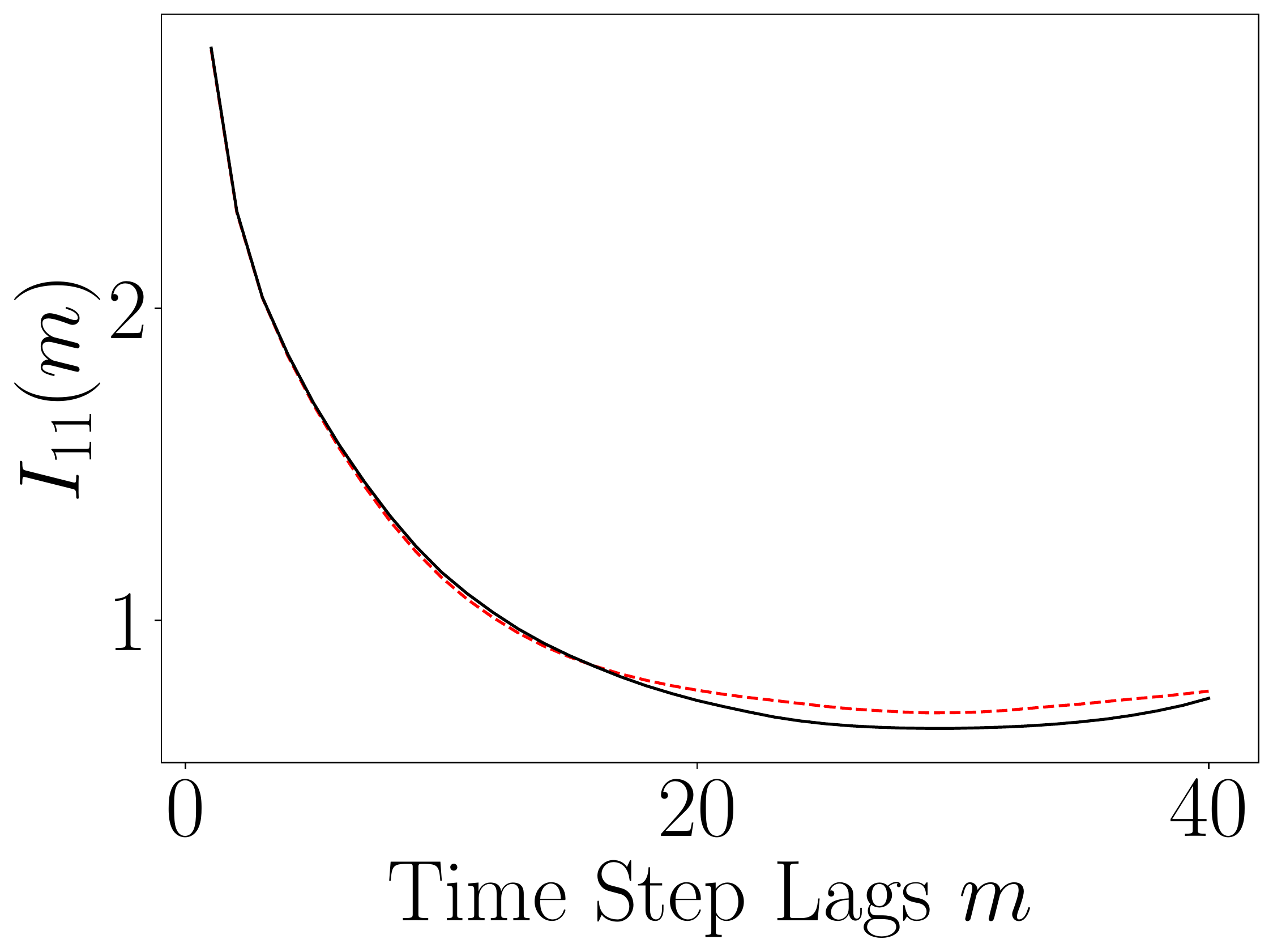} & \includegraphics[width=2.25in, height = 1.75in]{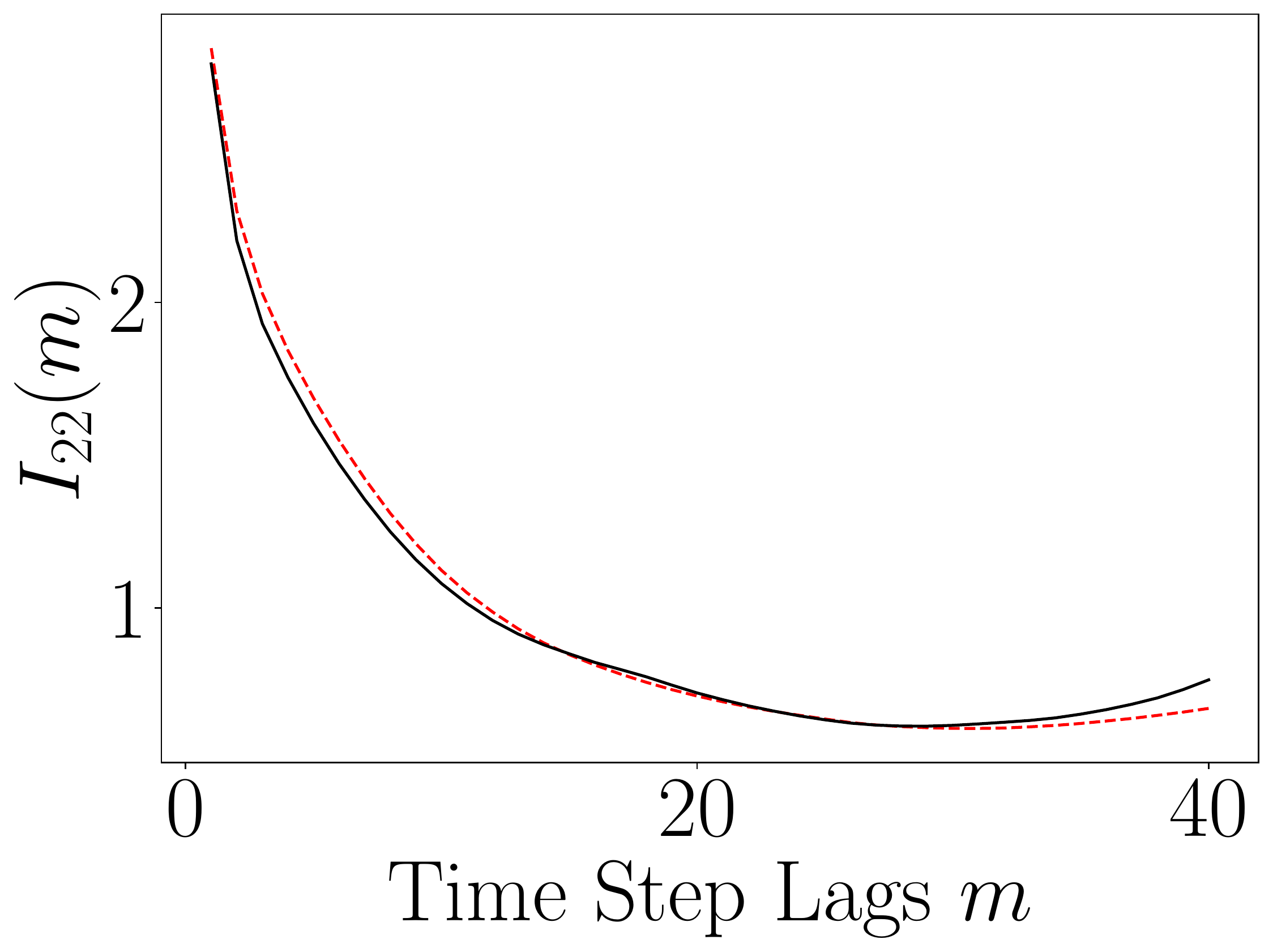} \\
(a)  &  (b) \\
 \includegraphics[width=2.25in, height = 1.75in]{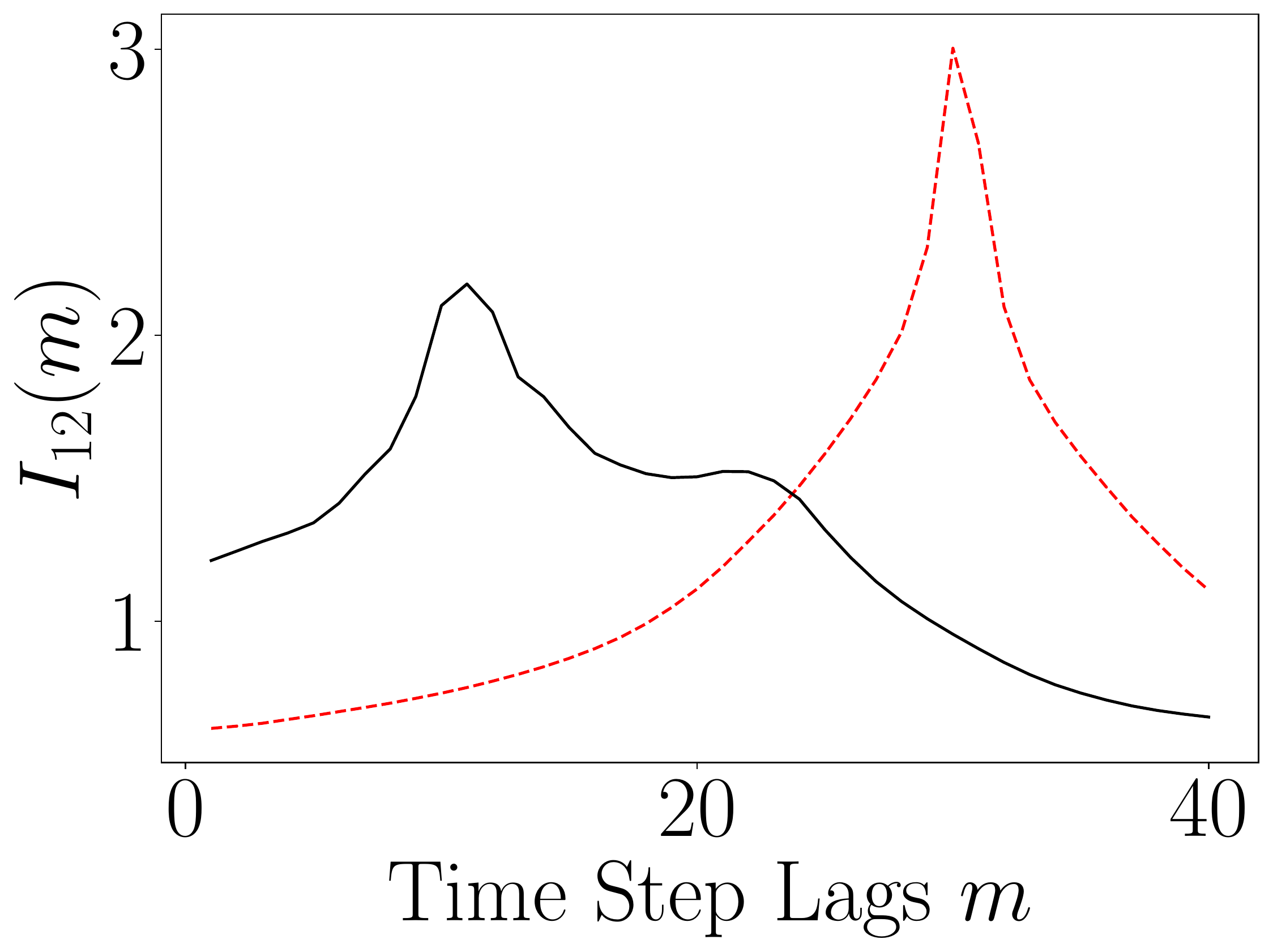} & \includegraphics[width=2.25in, height = 1.75in]{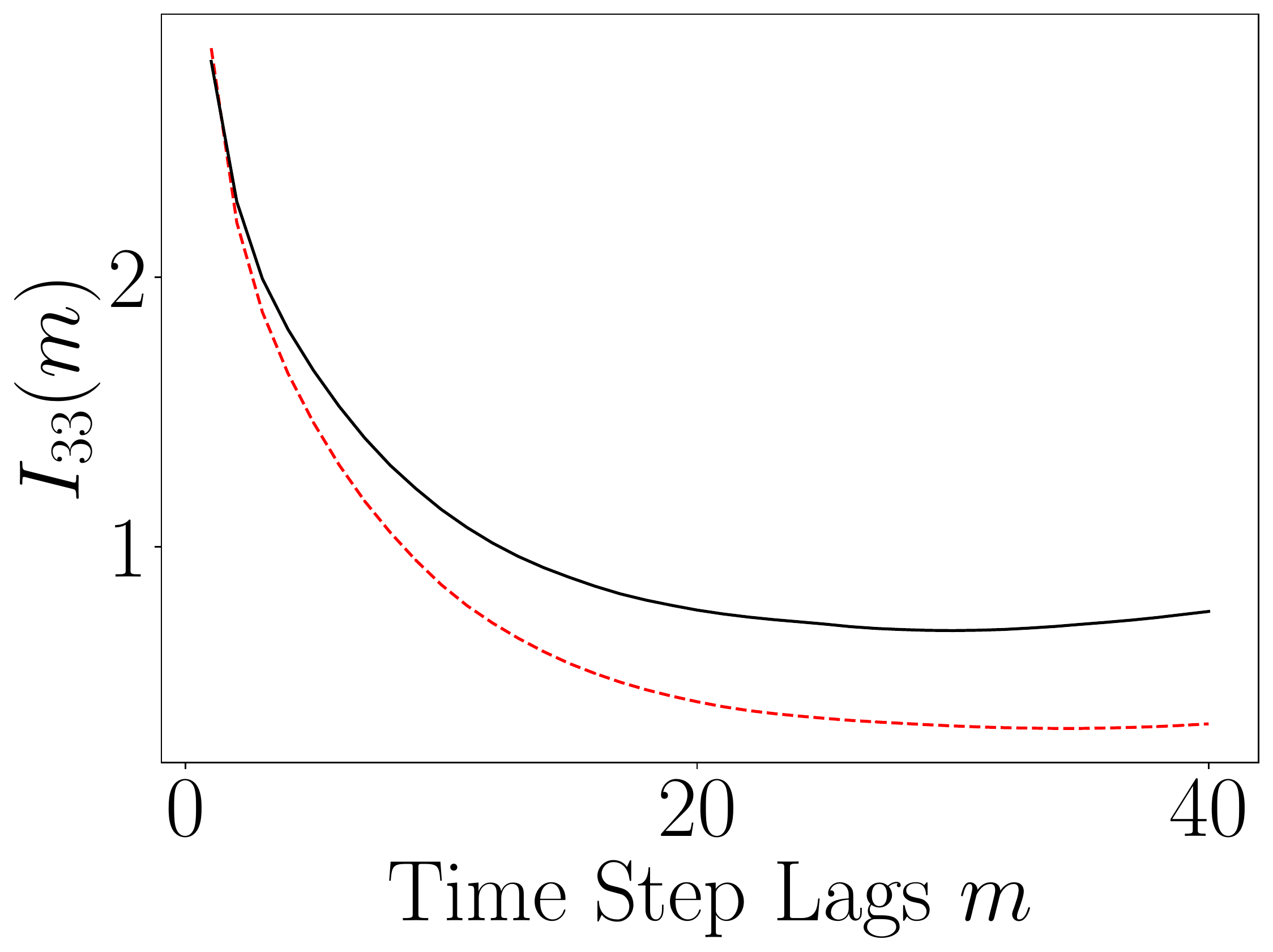} \\
(c) & (d)\\
\includegraphics[width=2.25in, height = 1.75in]{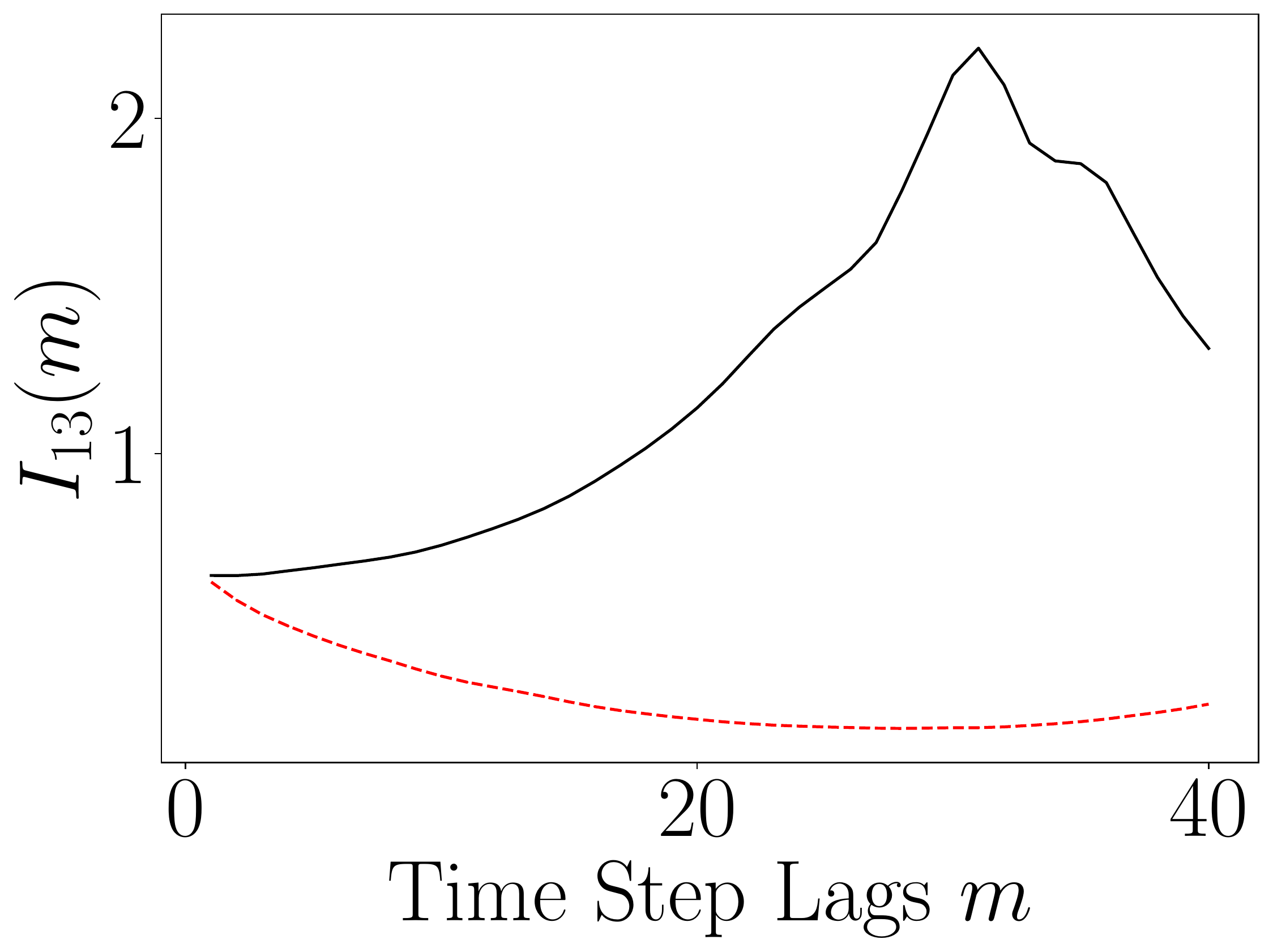} & \includegraphics[width=2.25in, height = 1.75in]{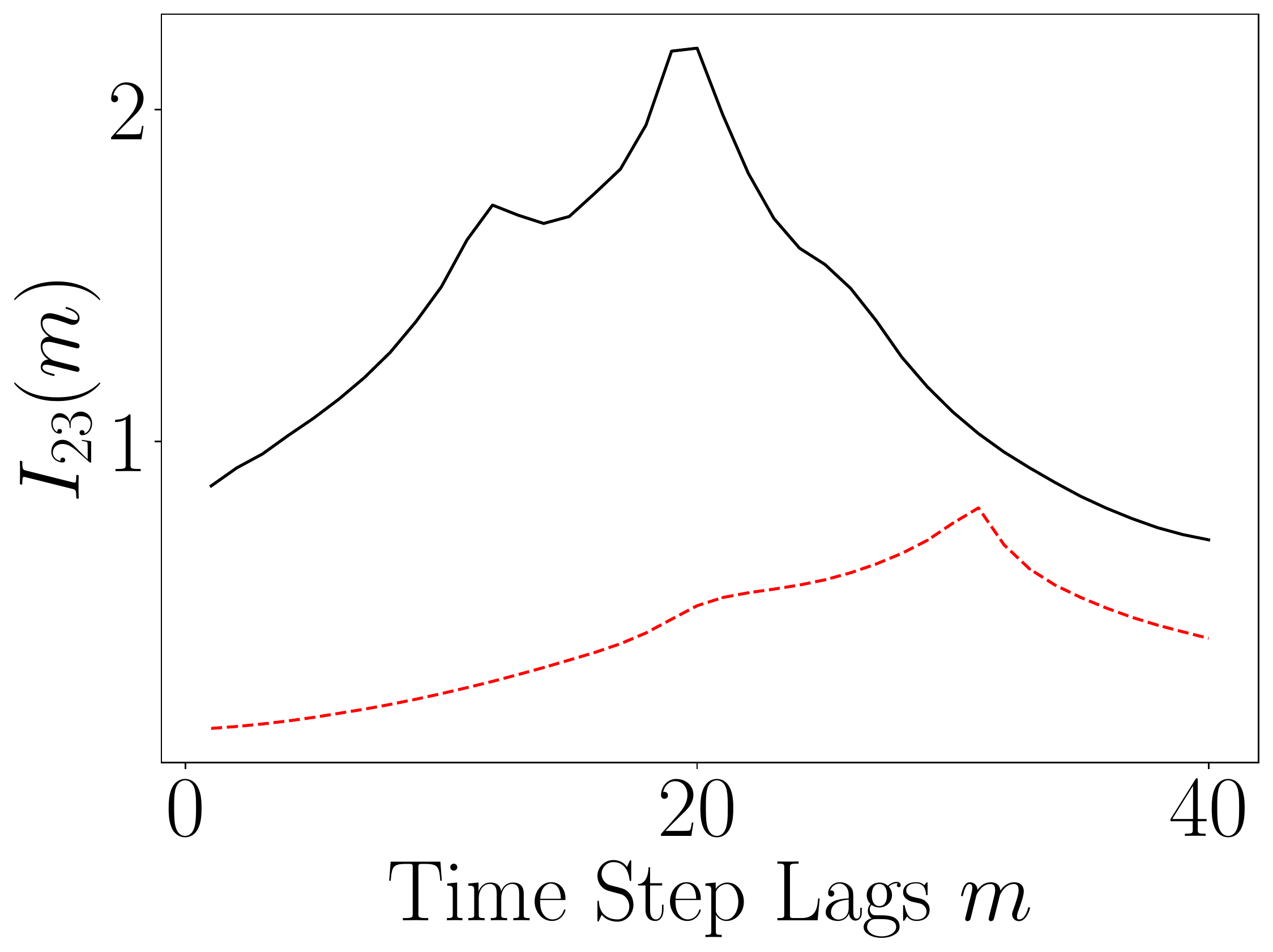}\\
(e) & (f)
\end{tabular}
\caption{For the Rossler system, plots of the ALSI $I_{nv}(m)$ for $(n,v)=(1,1)$ (a), $(n,v)=(2,2)$ (b), $(n,v)=(1,2)$ (c), $(n,v)=(3,3)$ (d), $(n,v)=(1,3)$ (e), and $(n,v)=(2,3)$ (f) for both the original (red/dash) and latent (black/solid) coordinates.}
\label{fig:micomps_rossler}
\end{figure}

When we examine the evolution over lags of the ALSI, we see in Figure \ref{fig:micomps_rossler} that the encoder is causing large and significant changes to the dynamics.  In particular, when we look at the plots of $I_{12}$  and $I_{23}$, we see that the sharp transients in the ALSI for the original coordinates is removed and the overall ALSI is relatively flattened in the latent coordinates.  This would seem to indicate that the slow/fast dichotomy in the Rossler dynamics is removed and so made more uniform.  Also of note though is $I_{13}$ which shows that the dependency between the $\tilde{y}_{1}$ and $\tilde{y}_{3}$ axes is enhanced relative to the coupling between $y_{1}$ and $y_{3}$ and that said dependency increases with lags.  

\section{Conclusion and Discussion}
In this work, we have developed a machine learning enhanced version of the HDMD and DLDMD which we call the DLHDMD.  We have shown that its performance is significantly better than just the HDMD, and when comparing against existing results in \cite{lago_dldmd} we see radical improvement over the DLDMD method for the Lorenz-63 system.  Likewise, we find that our method is successful across several challenging chaotic dynamical systems varying in dynamical features and size.  Thus, we have an accurate parallel approach fitting within the larger framework of Koopman operator based methods.  Moreover, we have a method which computes Koopman modes globally.  Finally, our analysis of the relative information dynamics across physical dimensions in the original and latent variables provides us a means of understanding the impact of the encoder network on the dynamics in line with modern thinking in machine learning as well as better pointing towards an understanding that the HDMD is enhanced by decreasing the relative statistical dependence across physical dimensions.  As explained in detail in the Introduction, there are of course a number of questions that remain to be addressed, and they will certainly be the subject of future research.  

\section{Acknowledgements}
C.W. Curtis would like to acknowledge the generous support of the Office of Naval Research and their Summer Research Faculty Program for providing the support of this project.  D.J. Alford-Lago acknowledges the support of the Naval Information Warfare Center.  E. Bollt was funded in part by the U.S. Army Research Office grant W911NF- 16-1-0081, by the U.S. Naval Research Office, the Defense Advanced Research Projects Agency, the U.S. Air Force Research Office STTR program, and the National Institutes of Health through the CRCNS.  We also wish to thank both anonymous reviewers for a number of insightful comments and questions which helped dramatically improve the paper.  

\section{Appendix}
\subsection{Models without HDMD for the Lorenz-63 System}
\begin{figure}[!h]
\centering
\begin{tabular}{c}
\includegraphics[width=4in, height=2.0in]{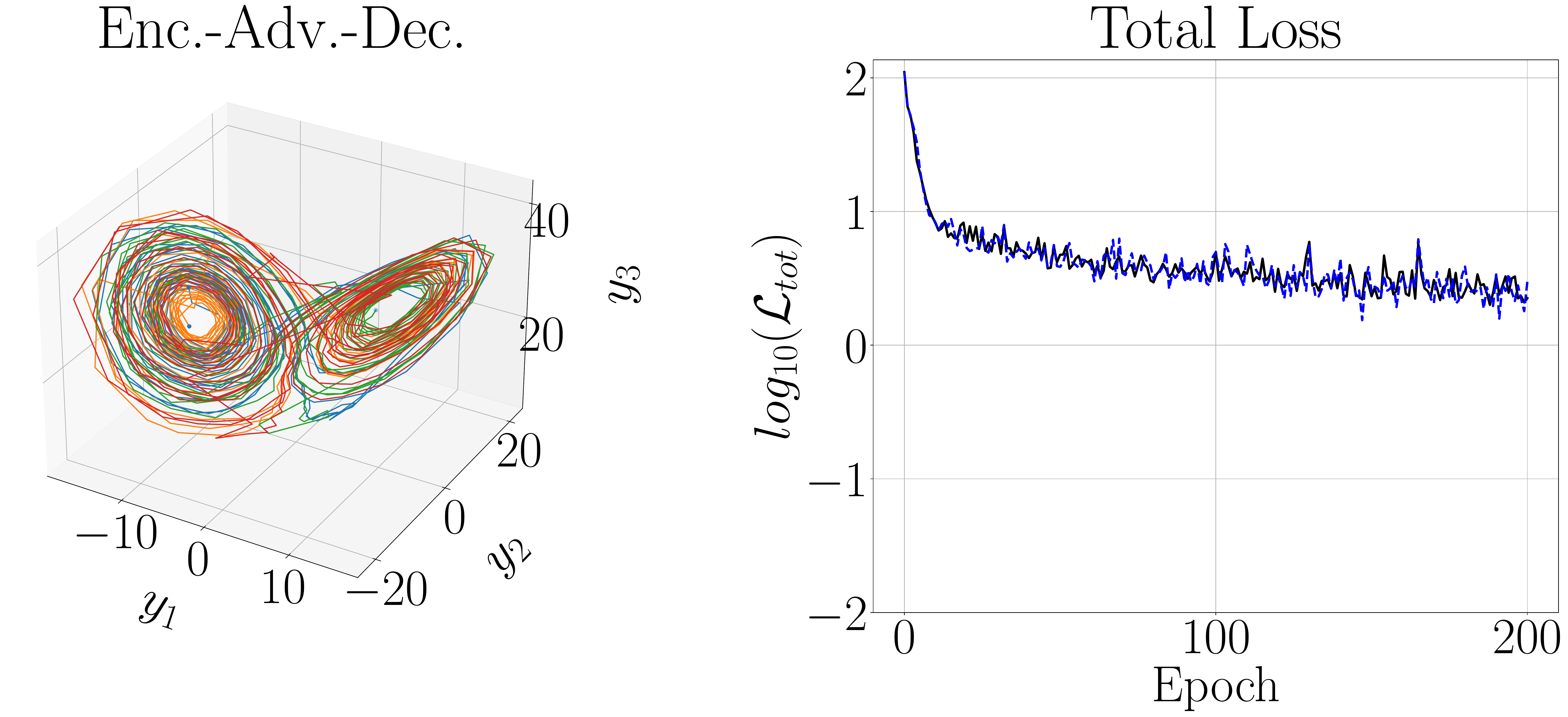}\\
$N_{e}=6, ~ \bar{N}_{ob}=1, ~ \text{Global EDMD}$\\
\\
\includegraphics[width=4in, height=2.0in]{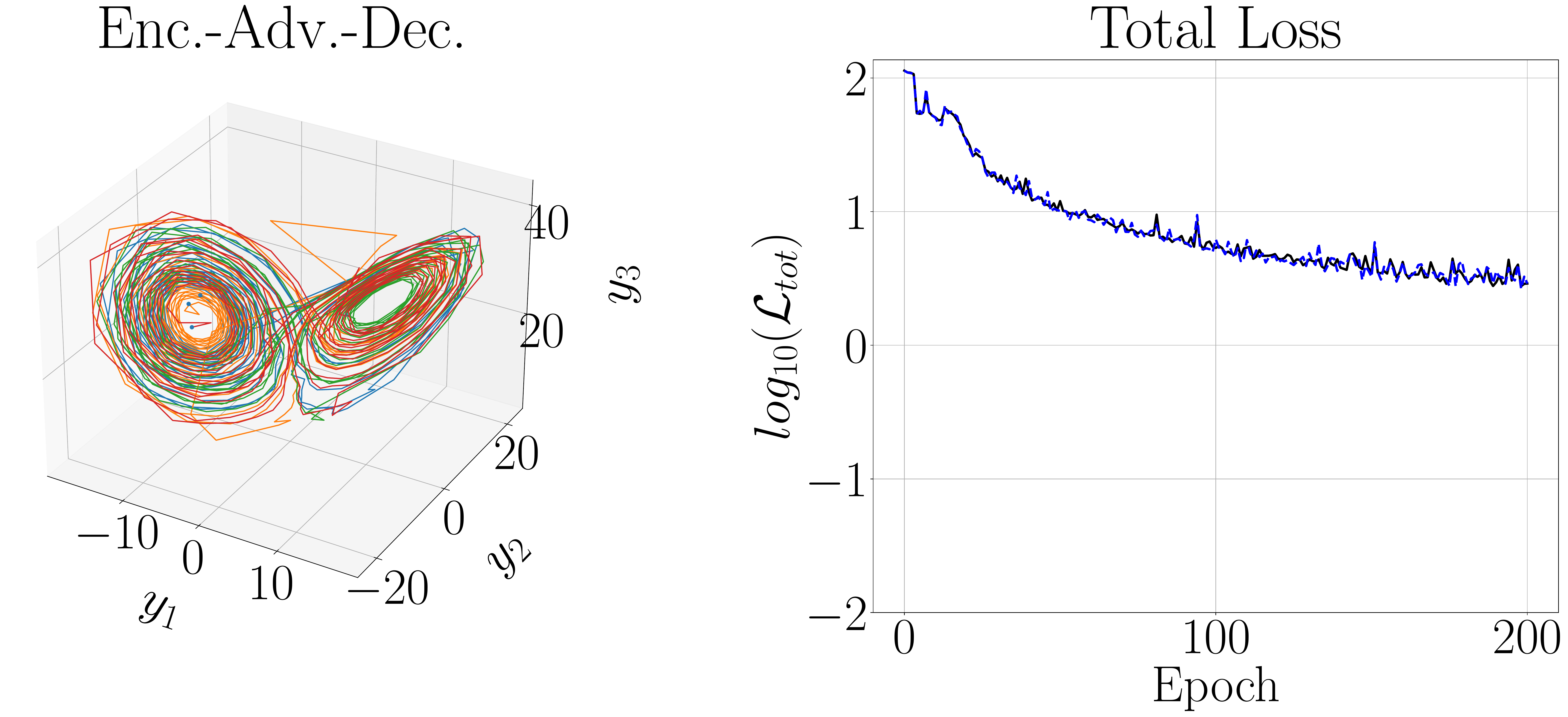}\\
$N_{e}=6, ~ \bar{N}_{ob}=1, ~ \text{Local EDMD}$\\
\end{tabular}
\caption{Results of the DLHDMD on the Lorenz-63 system after 200 epochs of training for $N_{e}=6$ and without updating the initial choice of $\bar{N}_{ob}=1$ using a global EDMD (top row) and a local EDMD (bottom row).  From left to right, we see the reconstruction generated by DLHDMD, i.e. $\mathcal{D} \left( \bar{{\bf K}}_{M}{\bf K}_{a}^{N_{st}}\mathbf{\Psi}_{-}\right)$, and the plot of $\mathcal{L}_{tot}$ over epochs.   The reconstruction is generated for times $1\leq t\leq t_{f}$ and forecasting is done for times $t_{f}\leq t \leq t_{f}+1$, where $t_{f}=20$.  Error plots are over both training (blue/dash) and testing (black/solid) data.  The reconstruction in the leftmost figure is generated over testing data.}
\label{fig:dldmd_nesix}
\end{figure}

Here, for the Lorenz-63 system, we explore the impact of fixing $\bar{N}_{ob}=1$, which is equivalent to setting $f_{r}=\infty$ and then removing any Hankel observables from our computations.  We also examine the difference between using a global EDMD versus a local one, at which point we are essentially using the DLDMD, i.e. local EDMD, algorithm presented in \cite{lago_dldmd}.  Parameter choices are kept the same for the sake of ready comparison.  For the DLDMD model, we also explored a range of embedding dimension choices where $N_{e}=4, ~5, ~ 6, ~ 7$, and $12$.  We found that $N_{e}=6$ gave the best results in tuning experiments.  That said, we emphasize that the worst results were found for $N_{e}=12$, and the differences between the choices in embedding dimension were not striking.  We can see the results of our final training and testing in Figure \ref{fig:dldmd_nesix}.  In either the global or local case, the results are far from desirable, especially compared to what we have thus far obtained through increasing $\bar{N}_{ob}$.  That said, while the rate of training is improved for the global EDMD model, in the end there is little difference in the final outcome between the local and global methods, and both clearly struggle to learn the dynamics, especially in comparison to our results using larger values of $\bar{N}_{ob}$, which is to say using Hankel DMD.  

\bibliography{hankel_dmd}
\bibliographystyle{unsrt}
\end{document}